\documentclass[runningheads]{llncs}
\usepackage{graphicx}
\usepackage{amsmath,amssymb} 
\usepackage{color}
\usepackage{adjustbox}
\usepackage{hhline}
\usepackage{caption}
\usepackage{placeins}
\usepackage[numbers]{natbib}
\usepackage{url}
\usepackage{epstopdf}
\begin{document}

\newcommand{\point}{
    \raise0.7ex\hbox{.}
    }


\pagestyle{headings}

\mainmatter

\title{Geodesic Distance Histogram Feature \\for Video Segmentation} 

\titlerunning{Geodesic Distance Histogram Feature for Video Segmentation} 

\authorrunning{Hieu Le\inst{1} \and Vu Nguyen\inst{1} \and Chen-Ping Yu\inst{2} \and Dimitris Samaras\inst{1}} 

\author{Hieu Le\inst{1} \and Vu Nguyen\inst{1} \and Chen-Ping Yu\inst{2} \and Dimitris Samaras\inst{1}} 
\institute{Stony Brook University\inst{1}, Harvard University\inst{2}} 

\maketitle

\begin{abstract}
This paper proposes a geodesic-distance-based feature that encodes global information for improved video segmentation algorithms. The feature is a joint histogram of intensity and geodesic distances, where the geodesic distances are computed as the shortest paths between superpixels via their boundaries. We also incorporate adaptive voting weights and spatial pyramid configurations to include spatial information into the geodesic histogram feature and show that this further improves results. The feature is generic and can be used as part of various algorithms. In experiments, we test the geodesic histogram feature by incorporating it into two existing video segmentation frameworks. This leads to significantly better performance in 3D video segmentation benchmarks on two datasets. 
\end{abstract}

\section{Introduction}
\label{sec:intro}

Video segmentation is an important pre-processing step for many high-level video applications such as action recognition \cite{Taralova2014}, scene understanding \cite{conf/iccv/JainCV13}, or 3D reconstruction \cite{Kundu2014}. A more compact representation not only reduces the subsequent processing space and time requirements, but also provides sets of visual segments that contain meaningful cues for higher-level computer vision tasks. However, generating supervoxels from videos is a significantly more difficult task than superpixel segmentation from images, due to the heavy computational cost and the extra temporal dimension.
Specifically, well delineated spatio-temporal video segments can be used for tracking bounded regions, foreground moving objects, or semantic understanding. For example, locating the movement of hands is helpful for gesture or action recognition, and separating foreground/background can pin-point the region-of-interest for detecting moving objects. Therefore, these spatio-temporal segments should be temporally consistent in order to be beneficial for these computer vision tasks.

For video segmentations that are initialized from superpixels, the main goal is to consider the connections between neighboring superpixels and to decide which ones belong to the same spatio-temporal cluster. The connections are usually represented as a spatio-temporal graph, where the nodes are the superpixels and the edges connect superpixels that are adjacent to each other. The edges are weighted based on the similarity distances between pairs of superpixels. Previous work \cite{Khoreva:7298697,Grundmann:5539893} proposed a variety of features corresponding to a wide range of low and mid-level image cues from superpixels. For example, the within-frame similarities were computed from boundary magnitude, color, texture, and shape, and the temporal connections were defined by the direction of optical flow or motion trajectories.  
Importantly, the aforementioned features that were used for video segmentation encode only local information, extracted from within each superpixel. One would expect improved performance when combining local and global features, if the appropriate global features per superpixel were extracted.

\begin{figure}[!t]
\begin{center}
\includegraphics[width=\textwidth]{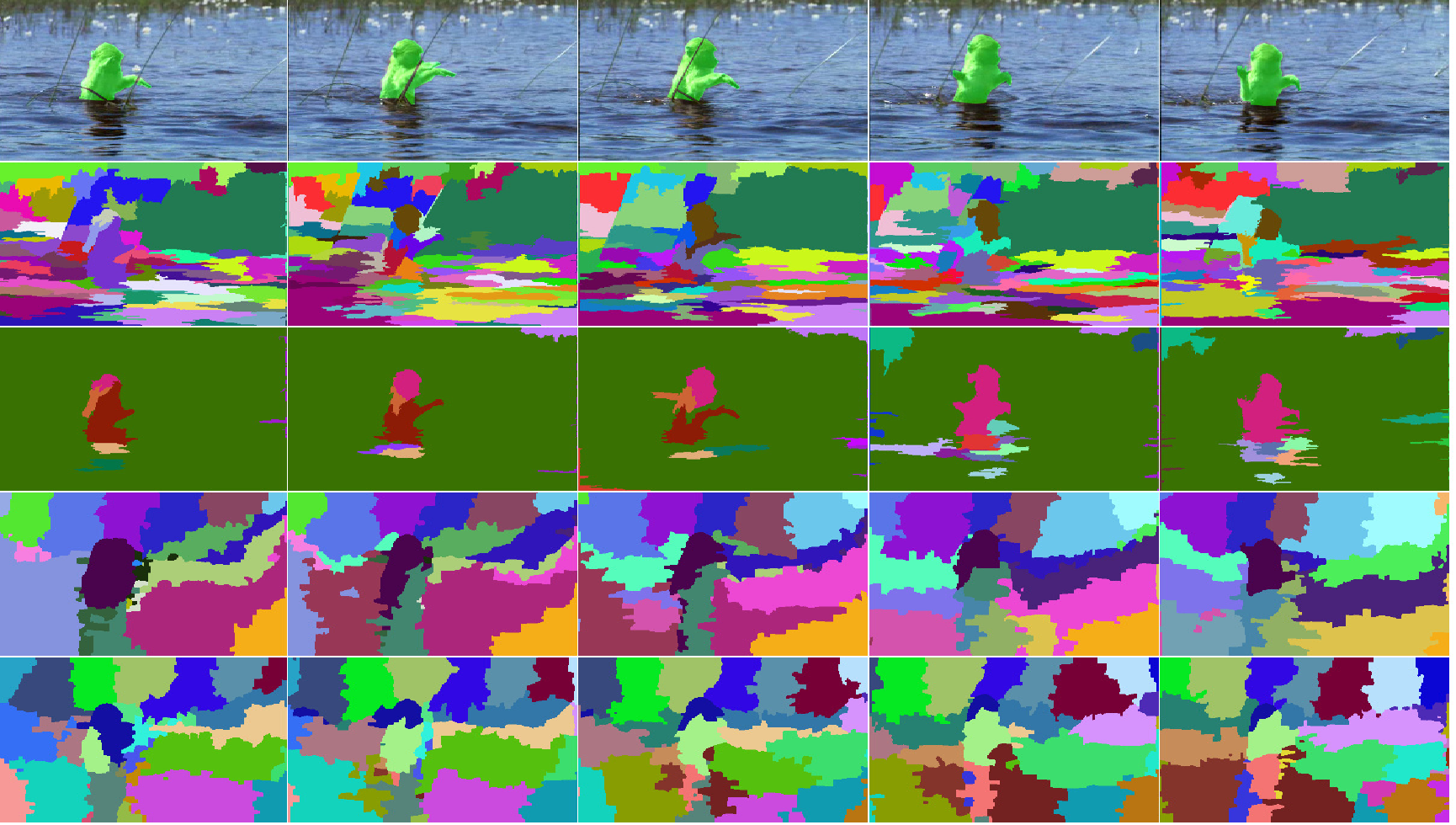}
\end{center}
\caption{The segmentation results on video ``monkey'' from Segtrack v2 dataset \cite{SegTrackv2_dataset}. Top row: original frames with superimposed ground-truth (green). Second row: segmentation results of the PGP algorithm (\cite{Yu_2015_ICCV}) using their four predefined features. Third row: result of PGP with our feature integrated. Fourth row: segmentation result of spectral clustering with the6 features proposed in \cite{Galasso2013}. Bottom row: segmentation result of spectral clustering with our feature integrated. Our results show better temporal consistency and less over-segmentation.}
\label{fig:preview}
\end{figure}

The geodesic distance has been shown to be effective for image segmentation problems \cite{Krähenbühl2014,Bai:4408931} but its applications in the video domain have been limited \cite{Wang:7298961,Bai:4408931,Price:5540079,geohis}. In this work, we propose a complete methodology for the use of geodesic distance histogram features in the video segmentation problem. The histogram feature describes the superpixel-of-interest by the distribution of the geodesic distances from it, to all other superpixels in the same frame. The representation compactly encodes global similarity relations between segments. Thus, we want to use per-frame geodesic distance information to associate superpixels both within and across frames. However, the nature of this global representation, poses several challenges that need to be addressed, in order to successfully use geodesic distance histograms for video segmentation:
\begin{itemize}
\item The feature needs to be robust across frames in order to perform useful superpixel association. That means if a superpixel has a unique representation in one frame, its representation in the next frame should be also unique, in order to facilitate matching.
\item For relatively small segments, their similar relationship to global context can dwarf distinctive neighborhood information, which might make them hard to differentiate.
\item The feature does not encode any spatial relationships between segments. Such relationships often offer constrains that allow otherwise similar segments to be distinguished from each other.
\end{itemize}

In this paper, we address these issues in order to derive a geodesic histogram feature that is appropriate for video segmentation tasks. In essence, we introduce the necessary local information in the global representation, in order to disambiguate associations across frames. For a given superpixel, we first extract the soft boundary map of the frame where it belongs, then we compute geodesic distances from the superpixel-of-interest to all other superpixels in the same frame using the boundary scores. If we were performing per frame segmentation, a 1D histogram of these scores would suffice \cite{Bai:4408931}. However, due to motion, this 1D histogram is not robust across frames. As observed  previously \cite{geohis}, a 2D joint histogram of intensity and geodesic distance is much more robust. To encode more spatial information into the feature, we compute multiple  geodesic histograms in a spatial pyramid \cite{1641019}. Finally, we weigh the bins with respect to their spatial distance from the superpixel-of-interest, in order to favor potentially discriminative neighborhood information. We show in experiments that when we add our complete geodesic histogram feature into existing frameworks, the resulting segmentations are greatly improved, especially in 3D segmentation accuracy and temporal consistency. The feature is also fast to compute, without increasing significantly processing time for the existing frameworks. The geodesic histogram features are added into two state-of-the-art video segmentation frameworks that are based on superpixel clustering, and tested on two popular datasets using standard 3D segmentation benchmarks.


The rest of paper is organized as follows: Section \ref{sec:rw} discusses related work. Section \ref{sec:ft} discusses the motivation, computation, and analysis of the proposed geodesic histogram features. Implementation details are described in Section \ref{sec:ID}. Section \ref{sec:exp} presents the experimental results. Section \ref{sec:cl}  concludes the paper and discusses other possible applications.



\section{Related Work}

\label{sec:rw}
 

Many video segmentation works propose diverse features to capture various kinds of information in order to estimate the similarity between the components of the video. Appearance can be represented by features based on color \cite{Grundmann:5539893,Cheng:6247744}, texture \cite{textons}, and soft boundaries \cite{Galasso_2014_CVPR}. Motion related features have also been utilized often, including short-term motion features based on optical flow \cite{conf/iccv/GalassoINC11,Tsai_2016_CVPR} and long-term motion features based on trajectories \cite{Bro10c,track_to_the_future,cPalou13,Brox:2010:OSL:1888150.1888173}. Superpixel shape is used to compute the similarities among superpixels across frames \cite{Cheng:6247744}. Some works discuss the choice of features to use \cite{Galasso2013} as well as the method to incorporate various kinds of features into affinity matrices \cite{Khoreva:7298697}.

Geodesic distances provide appearance-based similarity estimates. Geodesic distances have been applied widely on segmentation related problems on images \cite{Krähenbühl2014,geohis,Bai:4408931}. A feature based on geodesic distance for matching images of deformed objects has been introduced in \cite{geohis}. The authors showed that the geodesic distance could be invariant to object deformations, by encoding pixels as color histograms on the surrounding pixels that have the same geodesic distances. The geodesic distance is also used to propose object segments on images \cite{Krähenbühl2014}, which is based on the correlation between the object boundary and the change in the geodesic distance transform. 
Several video segmentation methods have employed geodesic distance for various purposes.
The salient object segmentation framework uses a geodesic distance in each frame to estimate the objectness of superpixels \cite{Wang:7298961} on a per frame basis. 
Further work further proposes a  spatio-temporal geodesic distance \cite{Bai:4408931} that extends image segmentation to video segmentation. However, the proposed spatio-temporal distance has to be constrained to be temporally non-decreasing to preserve the metric property, thus limiting the robustness of the method.

In this paper, we propose a feature based on geodesic distance to estimate the similarity between the superpixels in the video. We consider the frame-wise distribution of the geodesic distances, i.e., the histogram of geodesic distances from each superpixel to all other superpixels in the same frame. This representation compactly encodes the relative similarity distances between the segment containing the superpixel-of-interest to all the  other segments on the frame. This global information therefore serves as a complement to the set of to the set of appearance, motion, and shape-based features which only encode information from the inner region of the superpixel-of-interest.

\section{Geodesic Distance Histogram Feature}
\label{sec:ft}
Given a frame of the video, let $X$ be the set of superpixels: $X=\{x_1,\ldots,x_n \}$. 
The frame is then represented by a non-negative, undirected graph $G=(X,E)$, where each value in $E$ is associated with a pair of neighboring superpixels in $X$, and the edge weight is computed as the boundary strength between the two superpixels. 
The geodesic distance between any two superpixels $x_i,x_j \in X$  is defined as the weight of the shortest path between the two superpixels in $G$.

Given a superpixel $x_i$ on a frame, the geodesic distance between $x_i$ and all other superpixels in the same frame is computed and pooled into a geodesic distance histogram. This histogram contains the global information of the frame with respect to $x_i$ in terms of geodesic distance distribution, and can be used for computing pair-wise superpixel similarity both within and across frames.




\label{sec:HF}
\subsection{1D Geodesic Distance Histogram.}
The simplest approach is to use an 1D histogram to describe the distribution of the geodesic distances, where a bin of the histogram represents the number of superpixels with a particular geodesic distance. This is similar to the concept of critical level sets \cite{Krähenbühl2014}, where each critical level defines a group of superpixels having their geodesic distances less than a certain threshold. Each bin of the histogram is then associated with a region in the image.

In order to keep our feature relatively constant across frames, the value of each bin should stay approximately the same. This means that the regions associated with each bin also remain relatively stable. Considering the superpixel (in red) shown in Fig. \ref{fig:main1}(a), two regions corresponding to the first two bins of the histogram are visualized in Fig. \ref{fig:main1}(b). The first bin collects the votes of all superpixels with the lowest geodesic distance interval, forming the region indicated by the leftmost arrow. However, the region corresponding to the second bin is the combination of superpixels from different semantic regions. The value of the second bin is therefore not robust since these regions could potentially move in different ways, and end up voting for different bins in  subsequent frames.

\subsection{2D Intensity-Geodesic Distance Histogram.}  
\begin{figure}[!t]
\begin{center}
\begin{tabular}{ccc}
\includegraphics[width=0.3\textwidth]{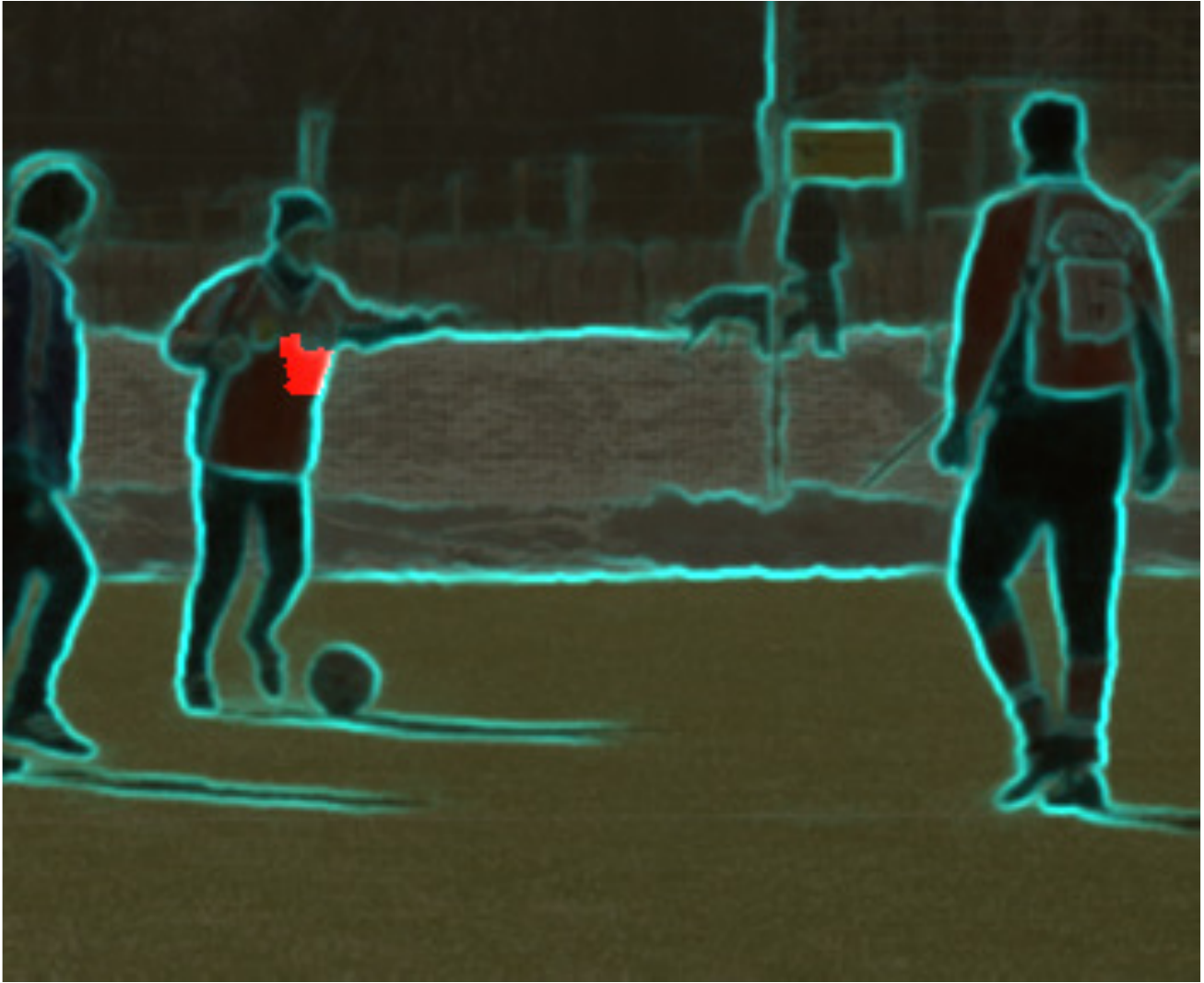}&
\includegraphics[width=0.3\textwidth]{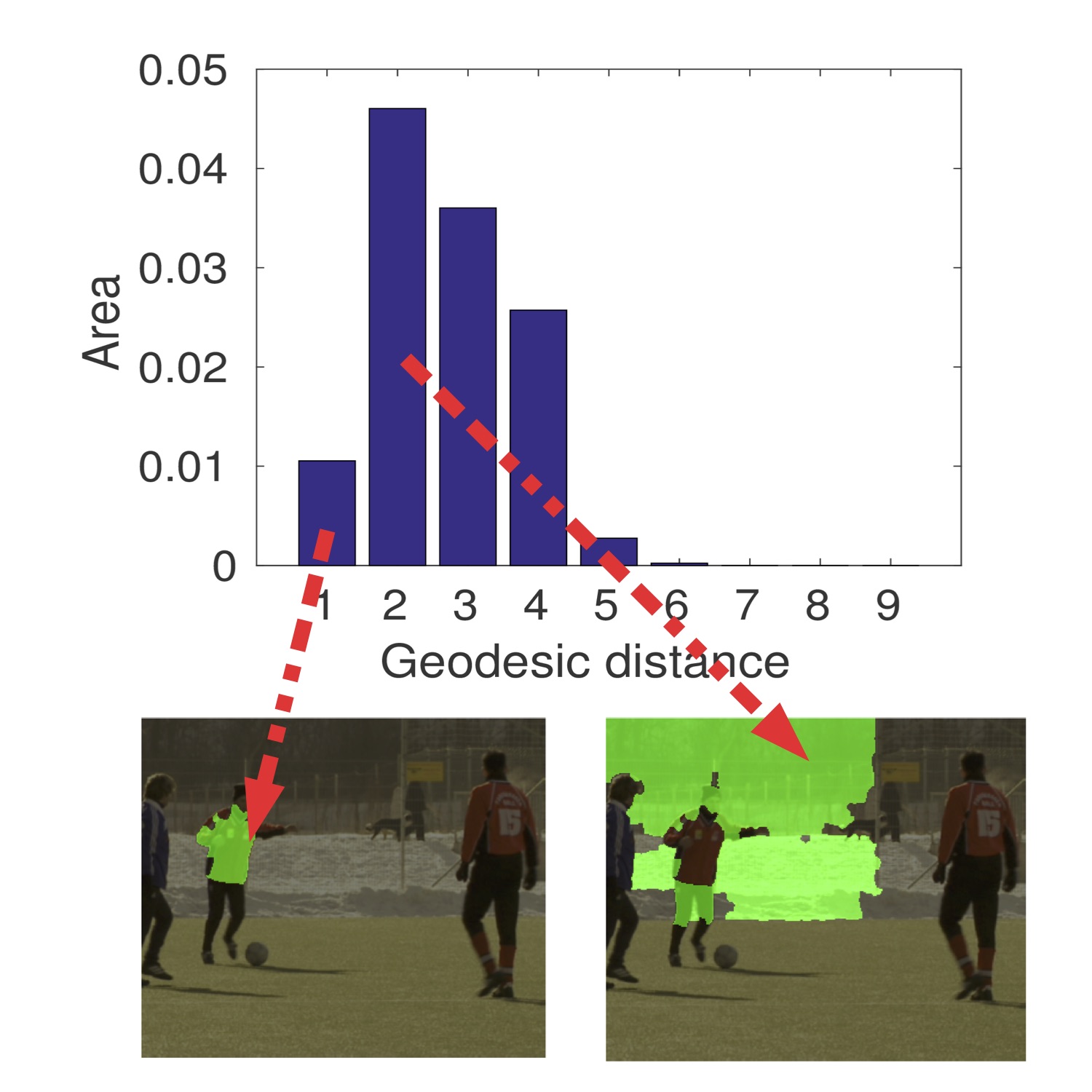}&
\includegraphics[width=0.3\textwidth]{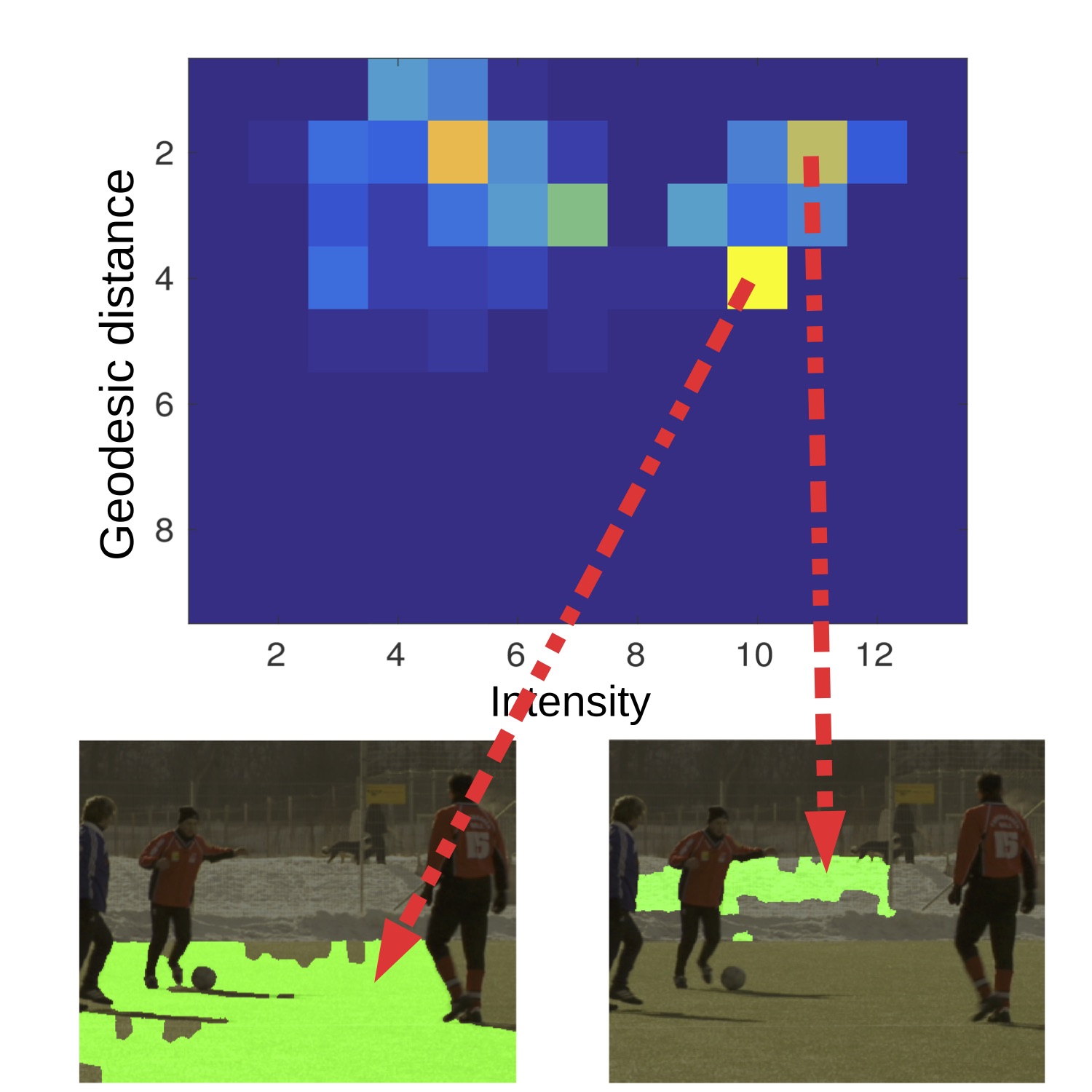}\\
(a)The superpixel-of-interest&(b)1D Histogram&(c)2D Histogram\\
\end{tabular}
\end{center}
\caption{The figure shows an example of 1D (geodesic distances) and 2D (intensity-geodesic distances) histogram features. (a): frame 1 of video ``soccer'' from Chen's Xiph.org dataset \cite{Chen_dataset}, with soft boundary scores highlighted, and a superpixel-of-interest marked in red. (b) and (c): the 1D and 2D histograms of the superpixel-of-interest, and the frame regions (green) that correspond to the selected bins and cells of the 1D and 2D histograms, respectively. (b) shows that the bins of the 1D histogram contain mixed information, while the cells in (c) contain regions that are more semantically homogeneous.}
\label{fig:main1}
\end{figure}
We incorporate the intensity feature as an additional cue  to complement the geodesic distance, on order to constrain bins to correspond to individual regions instead of disparate groups of regions. Thus the histogram becomes a 2D table where each cell is voted for by the superpixels that have a particular pair of geodesic distance and intensity. The joint distribution of  intensity-geodesic distance was originally proposed in \cite{geohis}, where the joint distribution was expected to be stable and informative under a wide range of deformations.

Fig. \ref{fig:main1}(c) visualizes the intensity-geodesic distance histogram of a superpixel-of-interest (shown in red in Fig. \ref{fig:main1}(a)). Notice that the second bin of the 1D histogram equals to the sum of all cells in the second row of the 2D histogram, and the region from the second bin in the 1D histogram is now separated into multiple smaller regions corresponding to these cells. This is a desired effect given that each of the cells in the 2D histogram  contains superpixels from the same semantic region as the 1D case. We also visualized the cell with the highest value in Fig. \ref{fig:main1}(c), which corresponds to the superpixels within the entire grass field. Such a region is likely to be stable across frames and remain connected. This implies that as long as the intermediate boundaries remain the same, these regions would still contribute to the same cells in the histogram.

To compute the similarity distance between two histograms, we can use the $\chi^2$ distance or the Earth Mover's Distance. Following \cite{geohis}, the  ${\chi}^2$ distance between two 2D histograms $H_p$ and $H_q$ with size $M \times N$ is  defined by:
\begin{equation}
{\chi}^2 (H_p,H_q) = {1\over  2} {\sum}_{k=1}^K {\sum}_{m=1}^M  {{[H_p (k,m) -H_q (k,m)]}^2 \over   H_p (k,m) +H_q (k,m)}
\end{equation}

The Earth Mover's Distance (EMD) is computed as the sum of the 1D EMDs at each intensity bin of the 2D histogram.

\begin{figure}[!t]
\begin{center}
\begin{adjustbox}{max width=0.9\textwidth}
\begin{tabular}{ccc}
\includegraphics[width=0.29\textwidth]{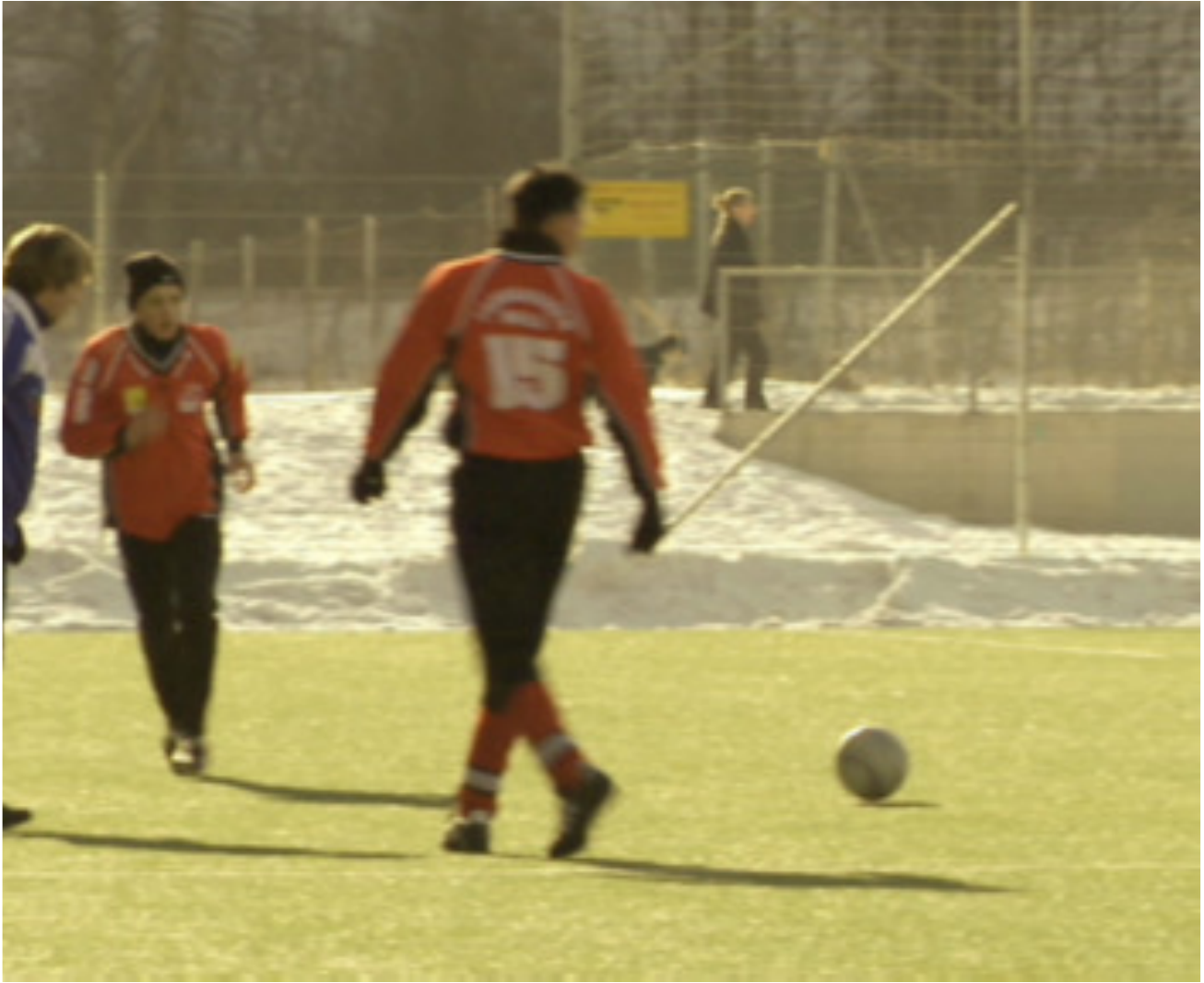}&
\includegraphics[width=0.3\textwidth]{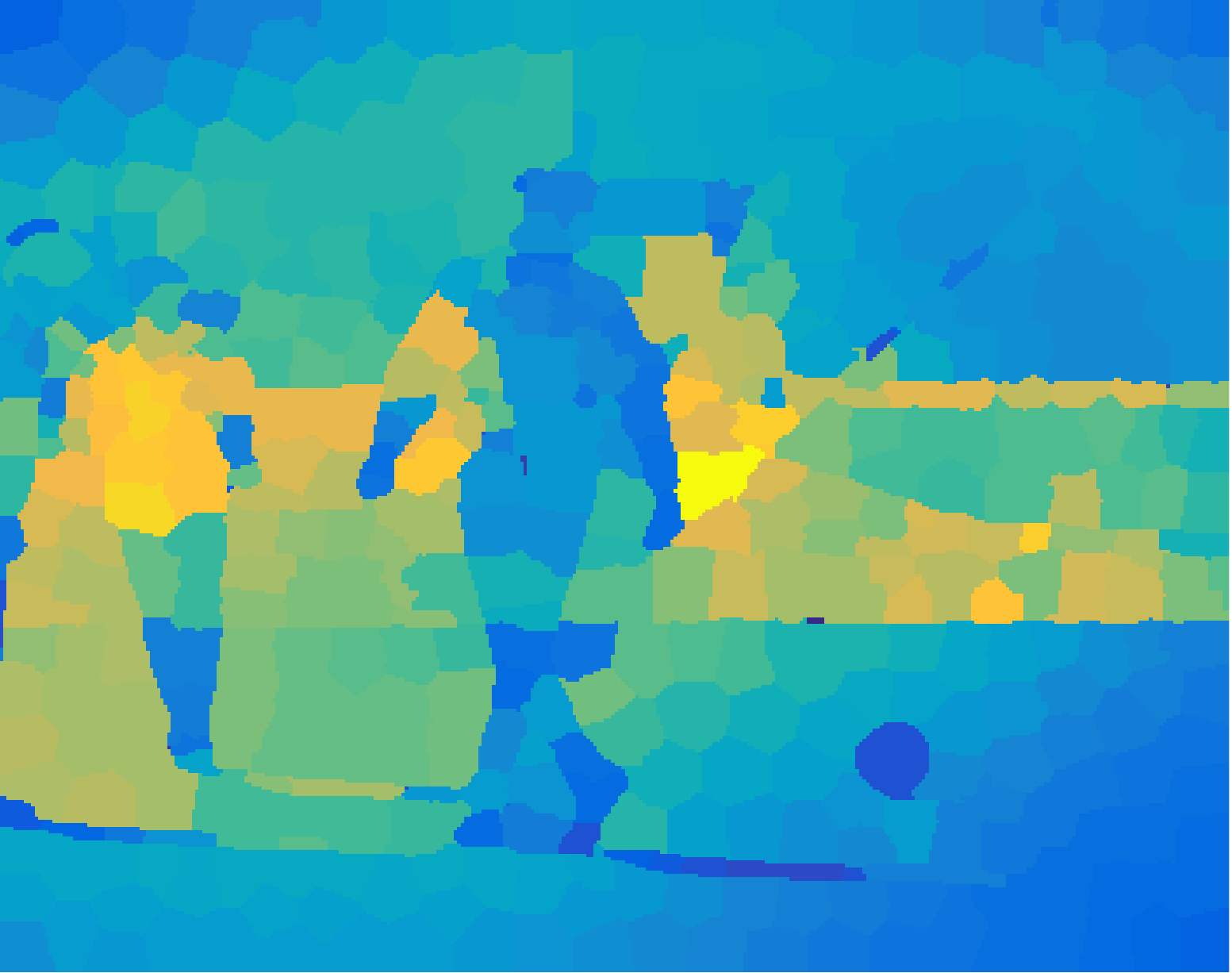}&
\includegraphics[width=0.3\textwidth]{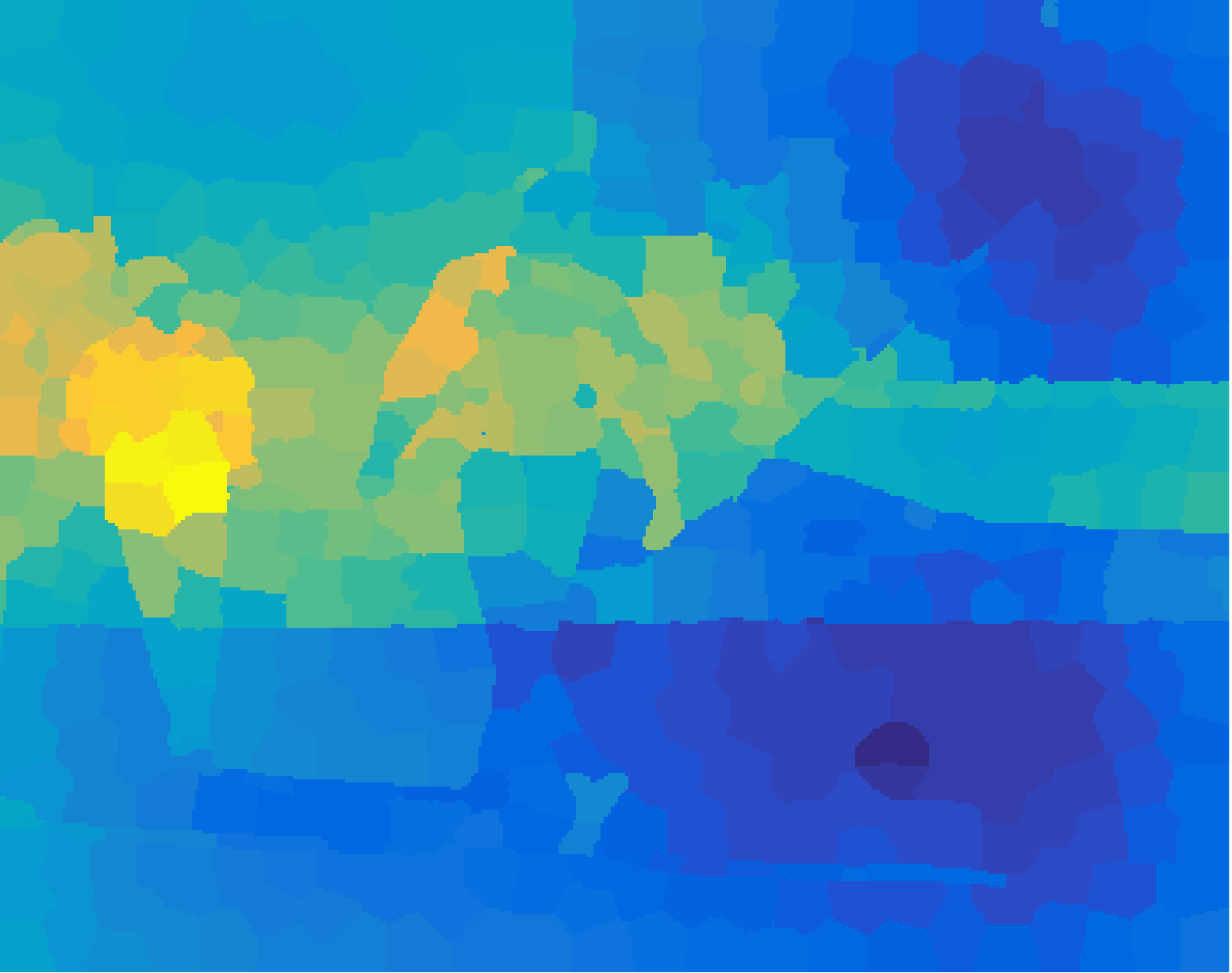}\\
(a)``soccer''-fr49&(b)1D across-frame&(c)2D across-frame\\
\end{tabular}
\end{adjustbox}
\end{center}
\caption{The figures visualize the similarities between the superpixel-of-interest in Fig. \ref{fig:main1}(a) on a later frame (frame 49) to all other superpixels. Warmer color represents higher similarity. (a): original frame. (b): the similarity map based on 1D geodesic distance histograms. (c): the similarity map based on 2D intensity-geodesic distance histograms. The figure shows that the 2D histogram is more robust than the 1D histogram for across-frame matching: there are multiple superpixels located in multiple regions that have similar 1D histograms with the superpixel-of-interest, while only the superpixels located within the upper-body region have the most similar 2D histograms.}
\label{fig:betweenfr}
\end{figure}

Fig. \ref{fig:betweenfr} visualizes the similarity values computed based on 1D and 2D feature histograms from the superpixel-of-interest in Fig. \ref{fig:main1}(a) on a later video frame. In the color scheme,  higher similarity is represented by the warmer color. The figure shows that the 1D histogram is less robust than the 2D histogram: there are multiple regions having similar 1D histograms with the superpixel-of-interest, and the superpixel with the highest 1D histogram similarity is in the background. In contrast, the superpixel with the highest similarity using the 2D histogram falls within the same upper-body region,  a desirable result.

\subsection{Spatial Information}
\label{sec:33}
Pooling methods such as histograms discard spatial information, such as image distance relationships or local neighborhood patterns. We encode spatial cues in two ways: 1) by embedding spatial distances into the voting weight of each superpixel, and 2) by adopting a commonly used spatial pyramid scheme \cite{1641019}.

\subsubsection{Spatial distance voting weight}
For a given superpixel $x$, its histogram feature is constructed by its intensity and geodesic distances to all other pixels in the same frame. To take the spatial location of these other superpixels into account, the geodesic distances are weighted by the spatial distance of those superpixels to $x$. In particular, the weighting of superpixel $y$ to the histogram bins of superpixel $x$ in frame $f$ is defined by:
\begin{equation}
weight_y = {|y| \over |f| }\times \exp({-\mu \times L_2(x,y) })
\end{equation}
where $|\cdot|$ is the area and $L_2(\cdot)$ is the Euclidean distance between two superpixels' center locations. 

The area component normalizes the influence of superpixels of different sizes.
The exponential ensures that nearby superpixels contribute more to the geodesic histogram of $x$. This is especially helpful for superpixels that belong to smaller segments, for which most other superpixels have large geodesic distances, that would dominate the histogram. Hence two small regions that are locally different would have very similar histograms. The parameter $\mu$ of the exponential controls the trade-off between global and local information. 

\subsubsection{Spatial pyramid histogram}
Inspired by the popularity of spatial pyramids \cite{1641019}, we incorporated the pyramid scheme into the construction of our feature histogram to encode more spatial information into the features. We implemented two scales of the spatial pyramid: 1x1 and 2x2 grids over a given frame. A histogram is extracted from each cell of the grid. Histograms from the same scale are concatenated.



\subsection{Implementation Details }
\label{sec:ID}
Our features are constructed from the intensity and boundary probability maps. For more robust boundary extraction, we also  experiment with two different boundary map methods: spatial edge maps using structured forests \cite{export:202540}, and motion boundary maps using the method proposed in \cite{Weinzaepfel_2015_CVPR}.  

Given the combined edge map and the superpixel graph, the geodesic distance feature for each superpixel is computed using Dijkstra's algorithm in $O(|X||E|log|X|)$, with the cost of a path being the accumulated boundary scores between one superpixel to another. 

We empirically set the intensity dimension of the feature histogram at 13 bins, and the geodesic dimension at 9 bins. 

\section{ Experiments}
\label{sec:exp}
In this section, we describe our experiments using the geodesic histogram features for video segmentation. We incorporated our features into two existing frameworks that are based on different clustering algorithms: spectral clustering \cite{Galasso2013} and parametric graph partitioning \cite{Yu_2015_ICCV}. Spectral clustering performs dimensionality reduction on an affinity matrix based on eigenvalues, while parametric graph partitioning directly performs the clustering on the superpixel graph by modeling $L_p$ affinity matrices probabilistically. Also, the method in \cite{Galasso2013} generates coarse-to-fine hierarchical segmentation results, while \cite{Yu_2015_ICCV} only outputs a single level of segmentation. 

The experiments were conducted on the Segtrack V2 \cite{SegTrackv2_dataset} and Chen's Xiph.org \cite{Chen_dataset} datasets, covering a wide range of scenarios for evaluating video segmentation algorithms. We evaluate our segmentation results using the metrics proposed in \cite{6247802}, including 3D Accuracy (AC), 3D Under-segmentation Error (UE), 3D Boundary Recall (BR), and 3D Boundary Precision (BP). All experiments were conducted with the exact same set of initial superpixels and other parameter settings.

\subsection{Video Segmentation Using Spectral Clustering}

We first evaluate the performance of the framework by adding our feature to spectral clustering  \cite{Galasso2013}. We use the same 6 features as  \cite{Galasso2013}: short term temporal, long term temporal, spatio temporal appearance, spatio temporal motion, across boundary appearance, and across boundary motion. The affinity matrix was computed by combining the 6 affinity matrices computed from each feature. We combined the original computed affinity matrix  with the geodesic histogram features in order to preserve the algorithm settings and superpixel configurations. The similarity distances based on our features were computed using the $\chi^2$ distance.

\begin{figure}[]
\begin{center}
\begin{adjustbox}{max width=0.9\textwidth }
\begin{tabular}{cc}
\includegraphics[width=0.5\textwidth,height=0.25\textheight]{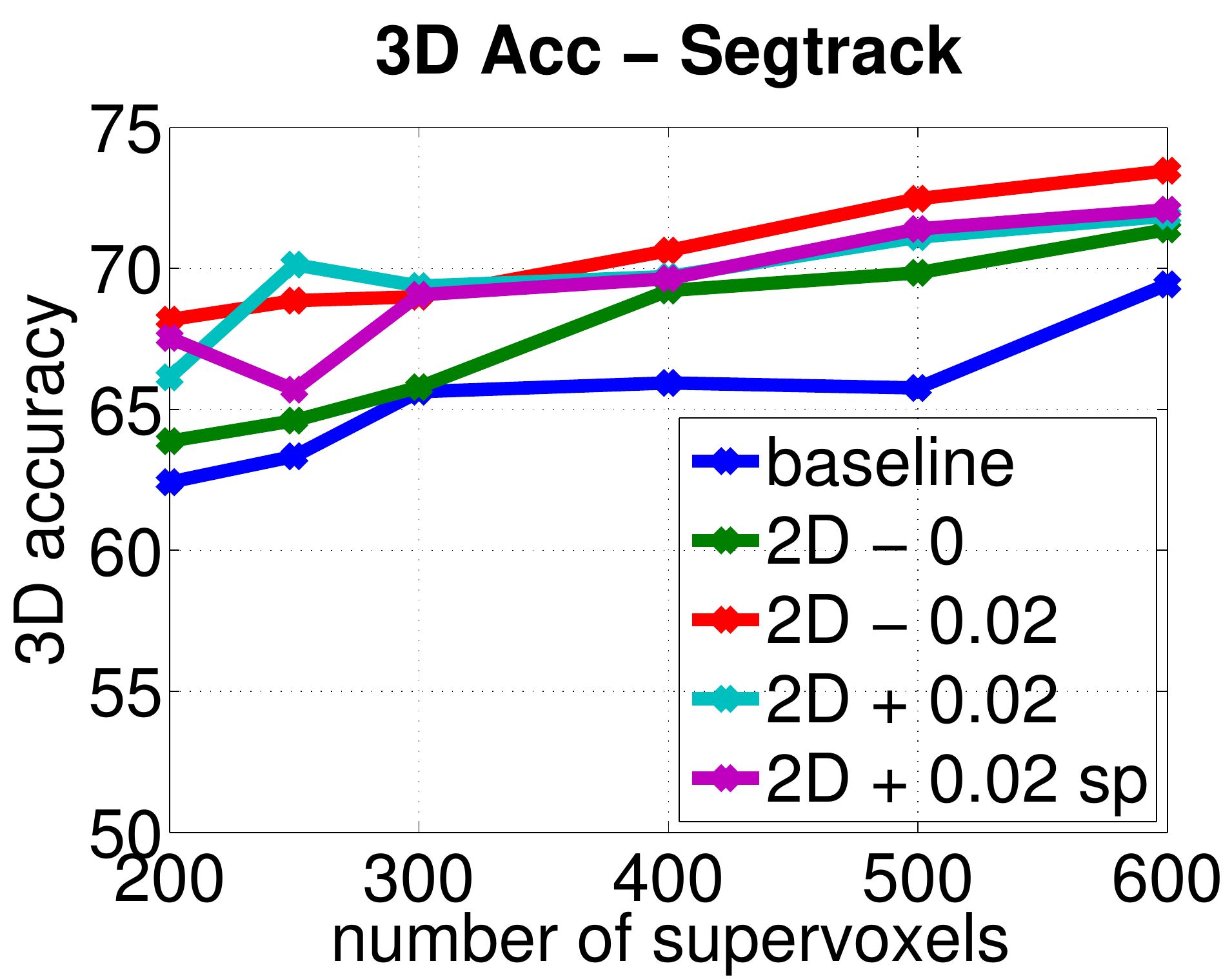}&
\includegraphics[width=0.5\textwidth,height=0.25\textheight]{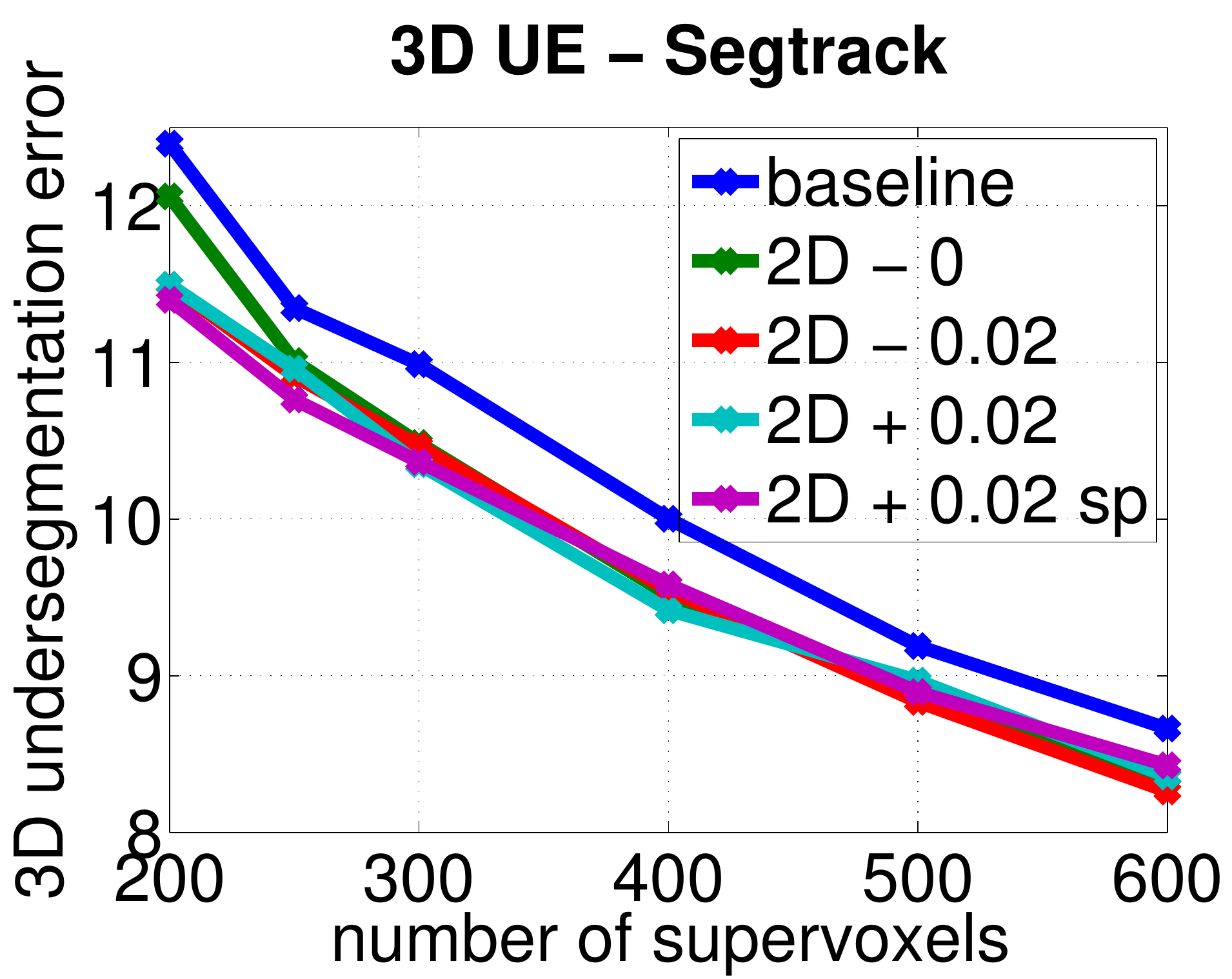}\\
\includegraphics[width=0.5\textwidth,height=0.25\textheight]{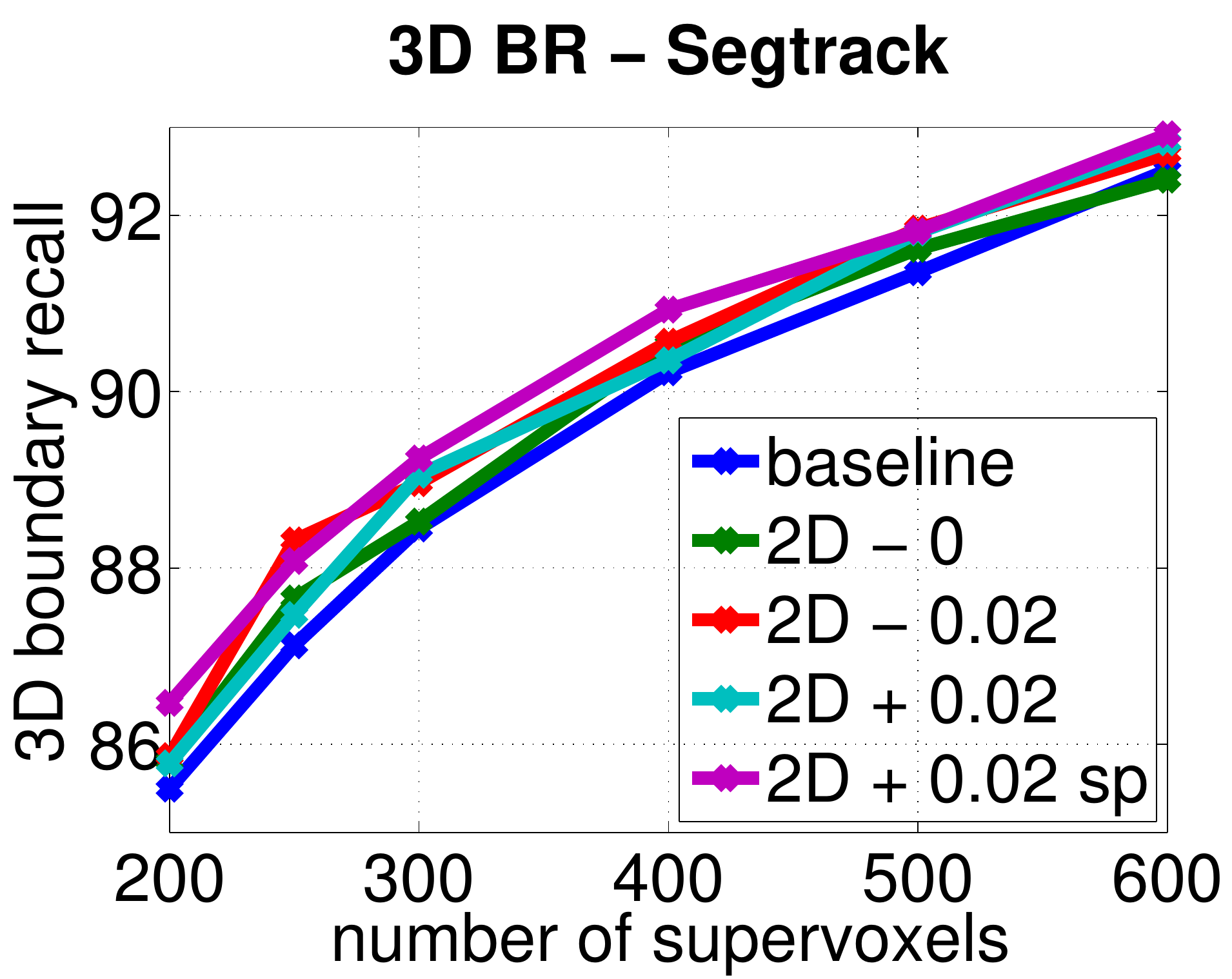}&
\includegraphics[width=0.5\textwidth,height=0.25\textheight]{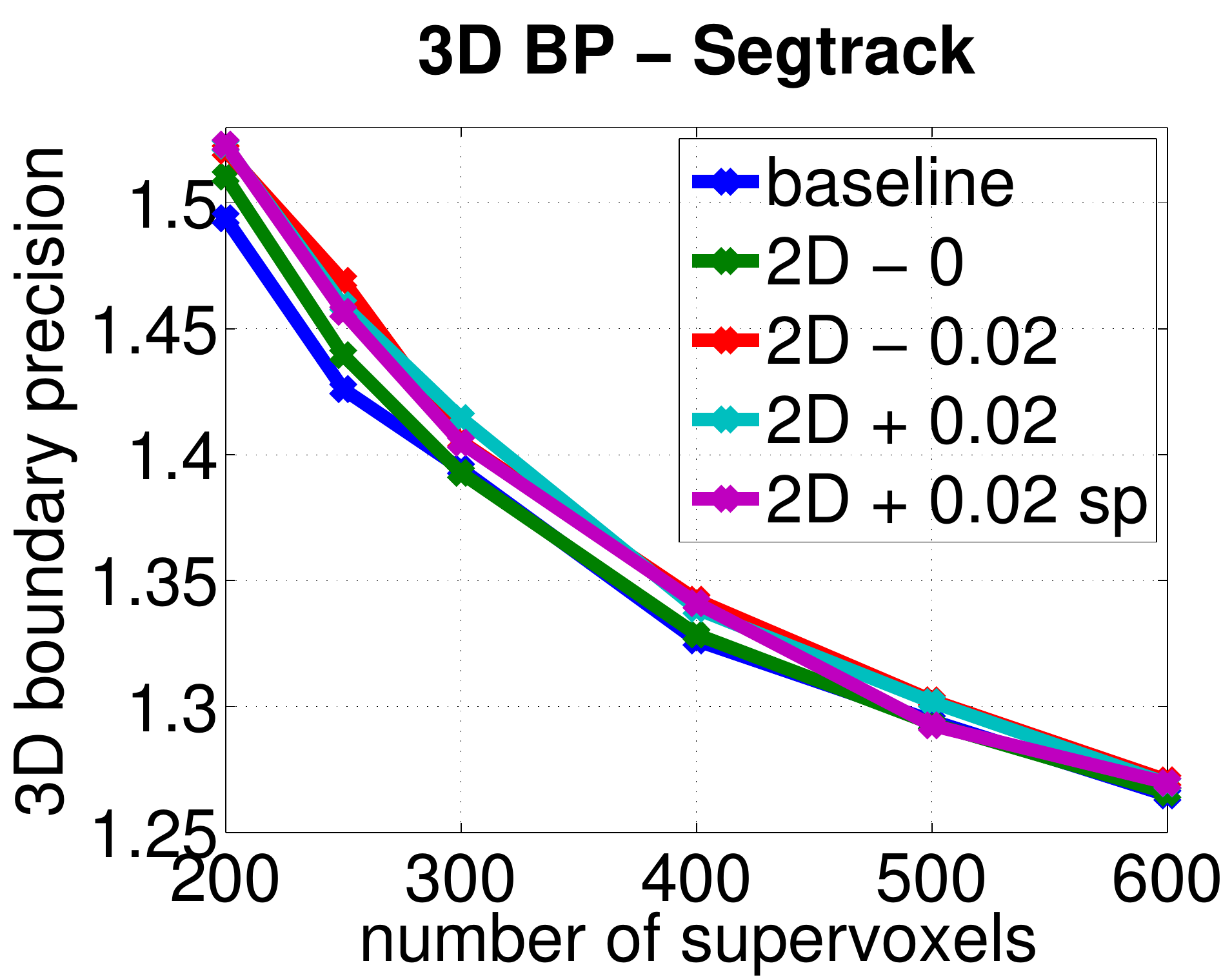}\\
\end{tabular}
\end{adjustbox}
\end{center}
\caption{Performance of spectral clustering (SC) \cite{Galasso2013} on the Segtrack v2 dataset, using four metrics: 3D Accuracy, 3D Under Segmentation Error, 3D Boundary Recall, and 3D Boundary Precision. For the 3D under-segmentation metric, the lower the error the better. For all the other metrics, the higher the score the better. -: using only spatial boundary edge. +: spatial boundary edge and motion boundary edge combined. 0: using spatial voting weight with $\mu =0$. 0.02: $\mu =0.02$. sp: with spatial pyramid. These plots show that the addition of our features result in major improvements on 3D Accuracy, and minor but consistent improvements on the three remaining metrics.}
\label{VSS_seg}
\end{figure}

Fig. \ref{VSS_seg} shows the evaluation results of spectral clustering with and without our feature on Segtrack v2 and Chen Xiph.org datasets. We tested four settings of our feature: \textbf{(i)} 2D histogram using only spatial edge maps to compute geodesic distances and without spatial distance voting weight (2D - 0), \textbf{(ii)} 2D histogram using spatial edge maps and spatial distance voting weight with $\mu = 0.02$ (2D - 0.02 ), \textbf{(iii)} 2D histogram using both spatial edge and motion boundary maps with $\mu = 0.02$ (2D + 0.02) and, \textbf{(iv)} 2D histograms with spatial pyramid (2D + 0.02 sp). Compared to the baseline, our feature significantly improved  segmentation performance. The improvement was most significant in 3D accuracy: increased by 5\% for Segtrack v2 and 10\% for Chen Xiph.org. For Segtrack v2 dataset, our feature was able to improve the segmentation results on all four metrics. For Chen Xiph.org dataset, the feature gave a strong boost to 3D accuracy and 3D boundary precision. For all settings tested, we noticed that motion boundary maps did not affect performance much. Given that motion boundary map generation requires optic flow computation, which can be time consuming, its omission might result in faster implementations. The spatial distance voting weights had a strong impact on the results and clearly improved segmentation.

\begin{figure}[!t]
\begin{center}
\begin{adjustbox}{max width=0.9\textwidth }
\begin{tabular}{cc}
\includegraphics[width=0.5\textwidth,height=0.25\textheight]{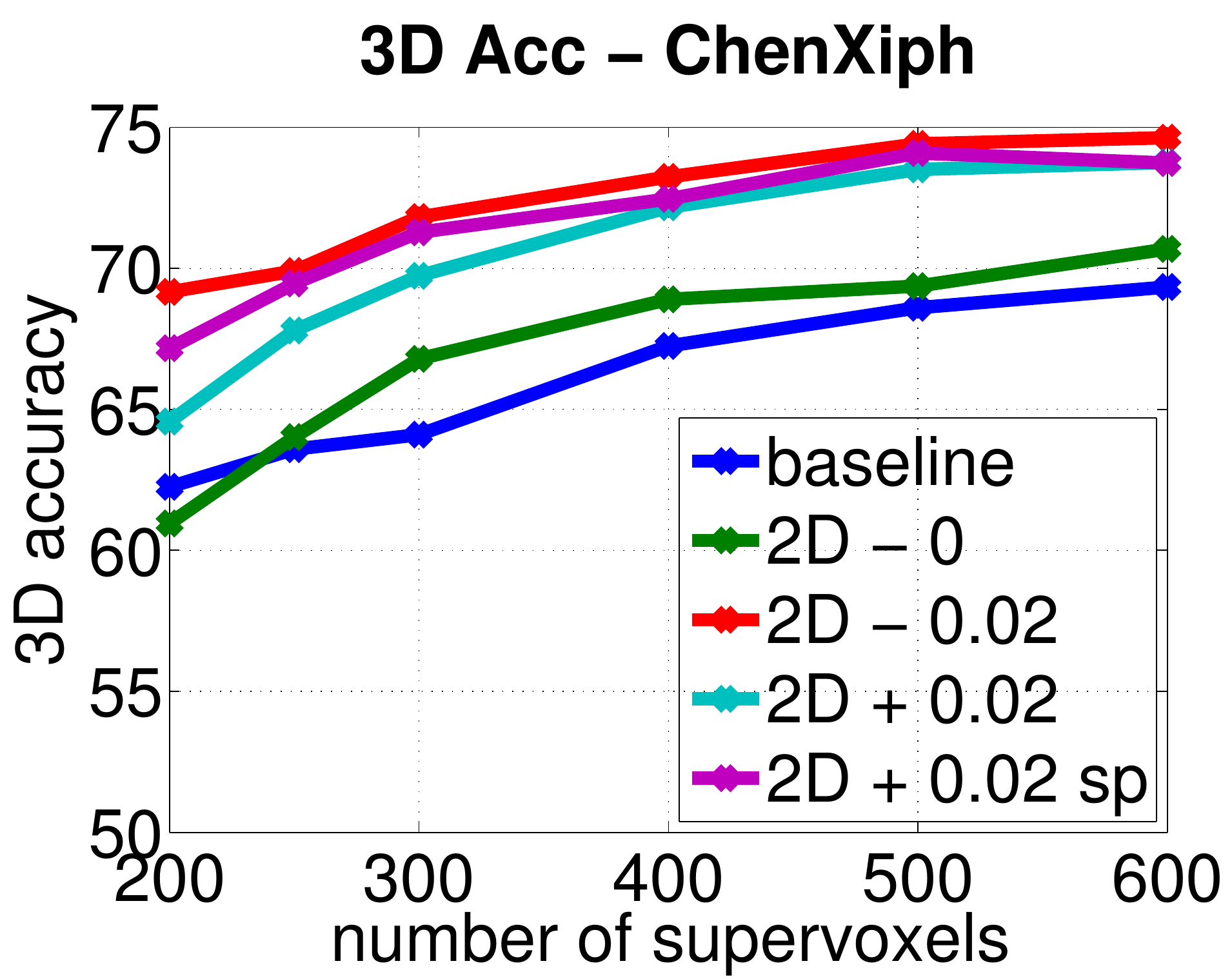}&
\includegraphics[width=0.5\textwidth,height=0.25\textheight]{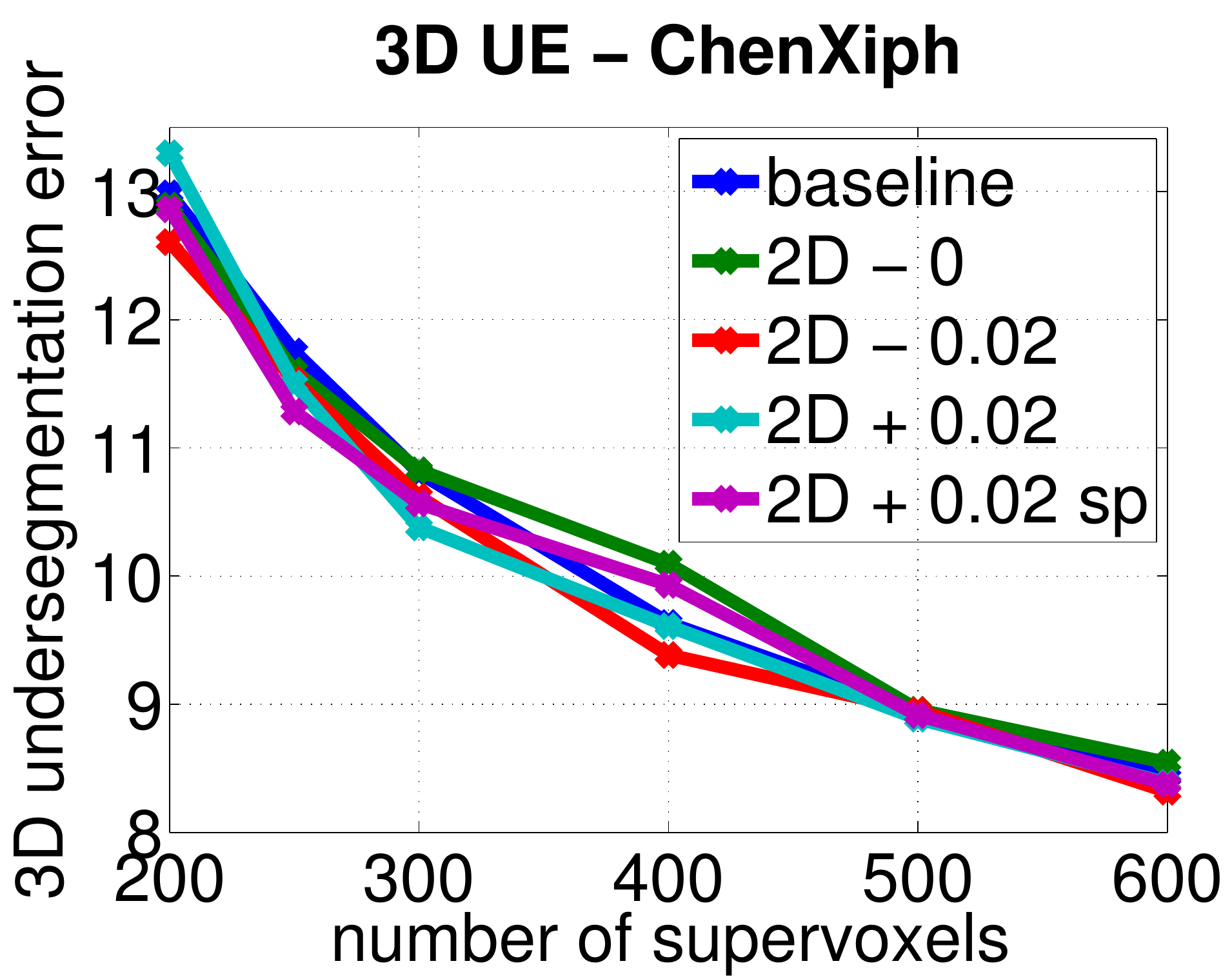}\\
\includegraphics[width=0.5\textwidth,height=0.25\textheight]{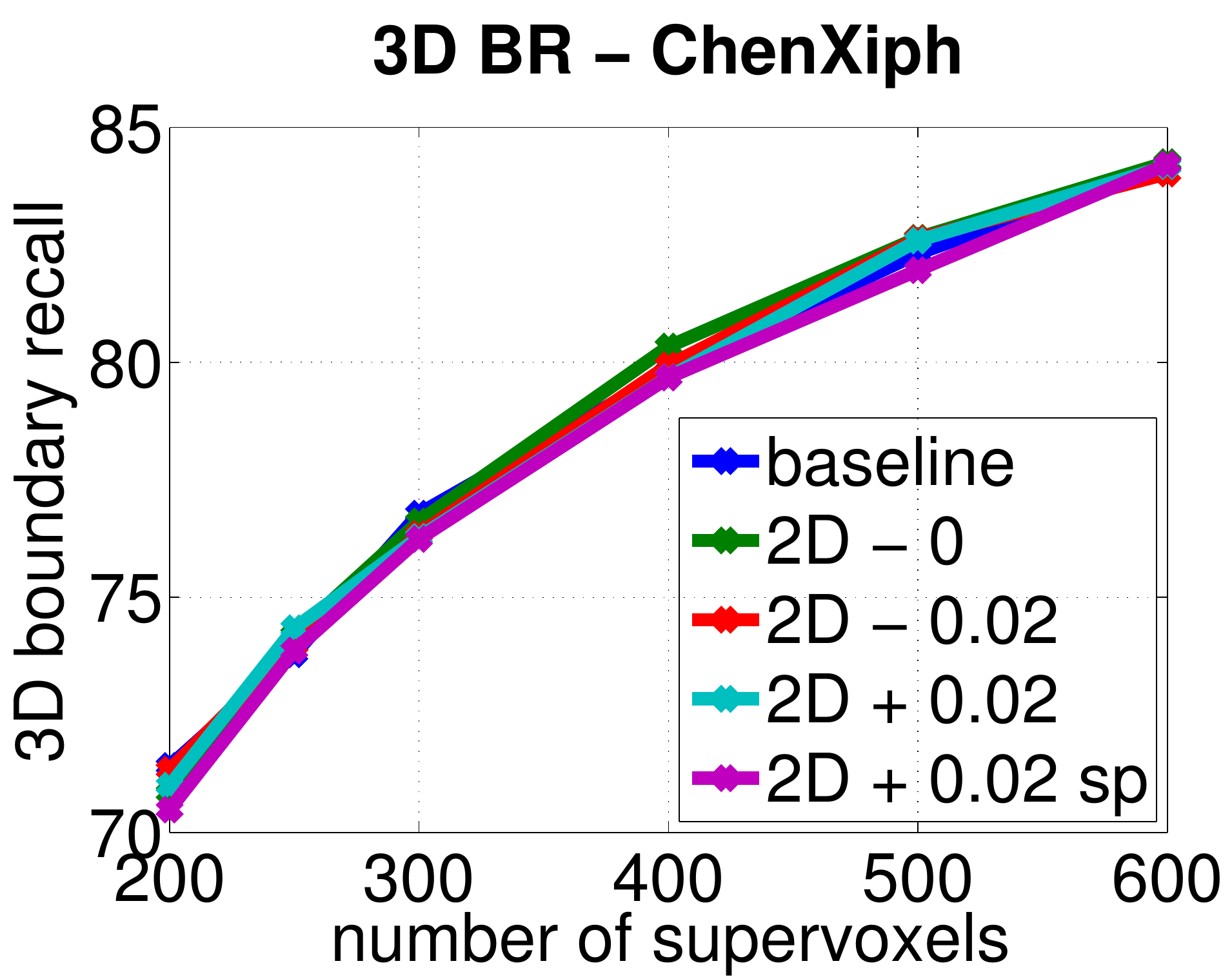}&
\includegraphics[width=0.5\textwidth,height=0.25\textheight]{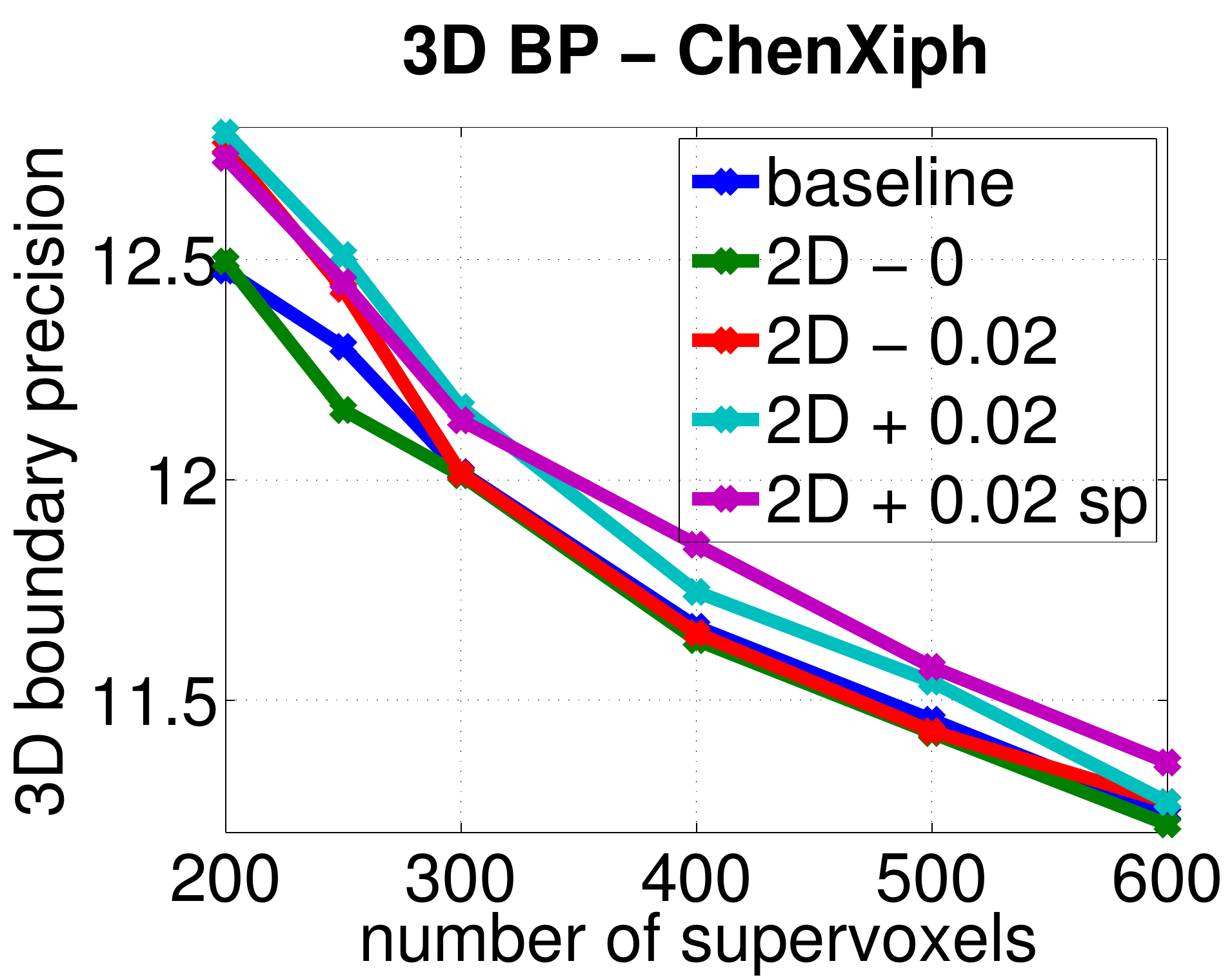}\\
\end{tabular}
\end{adjustbox}
\end{center}
\caption{Performance of spectral clustering (SC) \cite{Galasso2013} on the Chen Xiph.org dataset, using four metrics: 3D Accuracy, 3D Under Segmentation Error, 3D Boundary Recall, and 3D Boundary Precision. For the 3D under-segmentation metric, the lower the error the better. For all the other metrics, the higher the score the better. -: using only spatial boundary edge. +: spatial boundary edge and motion boundary edge combined. 0: using spatial voting weight with $\mu =0$. 0.02: $\mu =0.02$. sp: with spatial pyramid. These plots show that the addition of our features result in major improvements on 3D Accuracy, and minor but consistent improvements on the three remaining metrics.}
\label{VSS_chen}
\end{figure}

\begin{figure}[h]
\begin{center}
\begin{adjustbox}{max width=0.9\textwidth }
\begin{tabular}{cc}
\textbf{SEGTRACK V2}&\textbf{\space CHEN XIPH.ORG}\\
\includegraphics[width=0.5\textwidth]{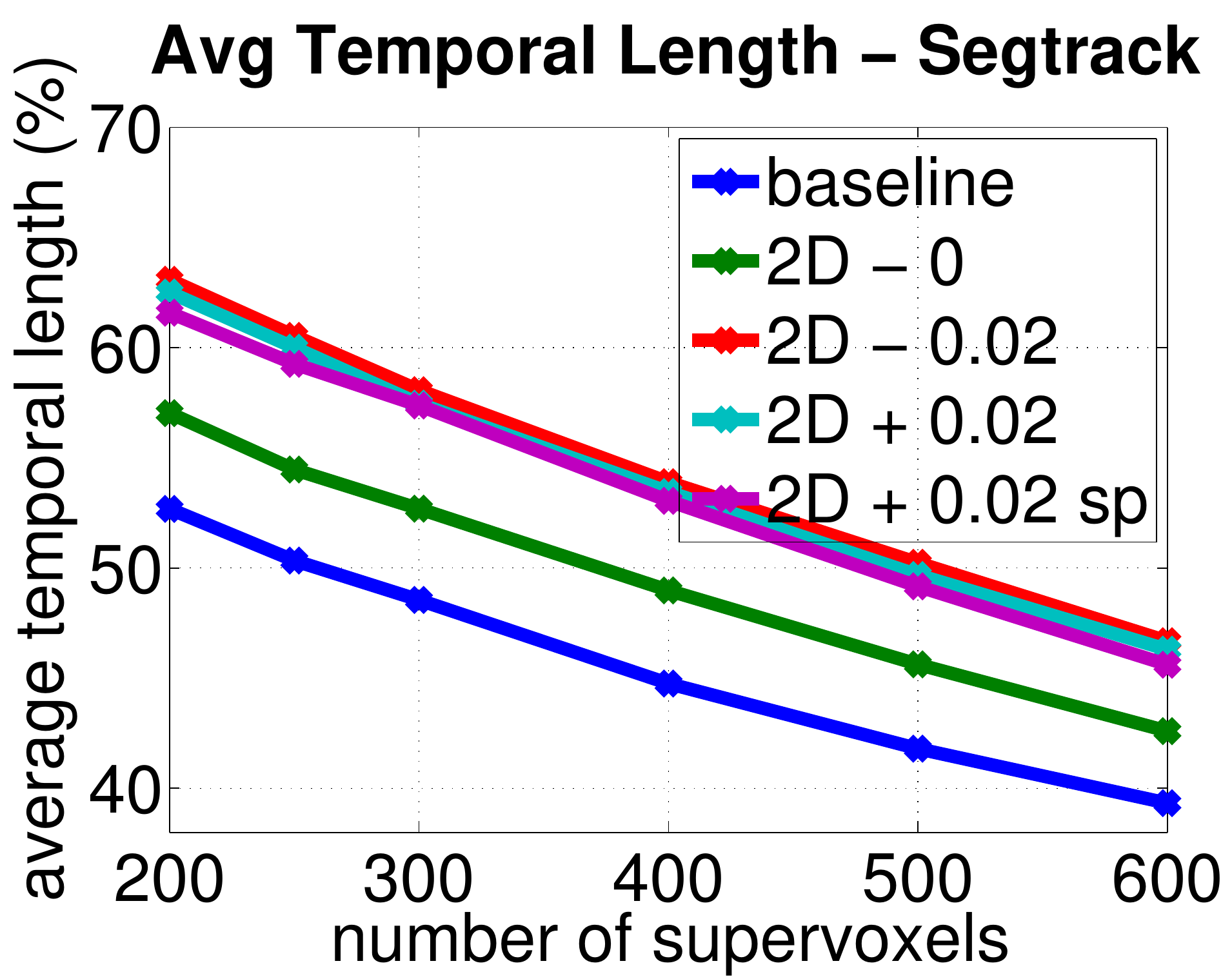}&
\includegraphics[width=0.5\textwidth]{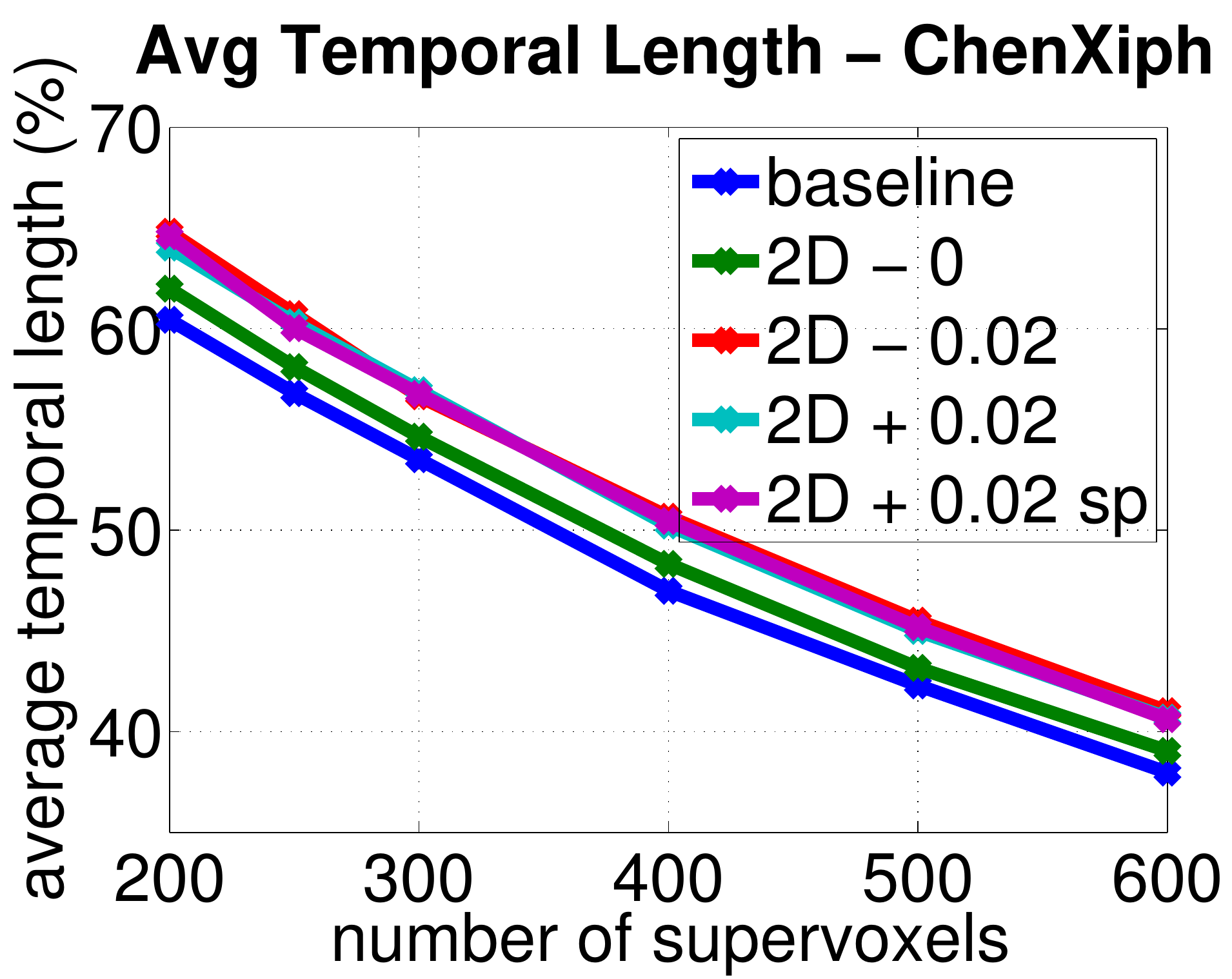}\\
\end{tabular}
\end{adjustbox}
\end{center}
\caption{ Average temporal length of supervoxels generated by spectral clustering (SC) \cite{Galasso2013} on the Segtrack v2 and Chen Xiph.org datasets. The results show significant improvements on the temporal consistency with the addition of our feature on Segtrack v2 dataset, and minor but consistent improvement on the Chen Xiph.org dataset.}
\label{tl}
\end{figure}

In addition to these improvements, Fig. \ref{tl} shows that the average temporal length of supervoxels consistently increased for all parameter settings of our feature by 10\% for Segtrack v2 dataset and 5\% for Chen Xiph.org dataset, showing that the segmentation results acquired better temporal consistency. Having both longer supervoxels and improved segmentation metrics indicate that our feature provides additional information for more reliable temporal consistency. This is significant, since connecting more corresponding superpixels temporally is a crucial and challenging part of the video segmentation task.

\begin{figure}[!t]
\begin{center}
\includegraphics[width=\textwidth]{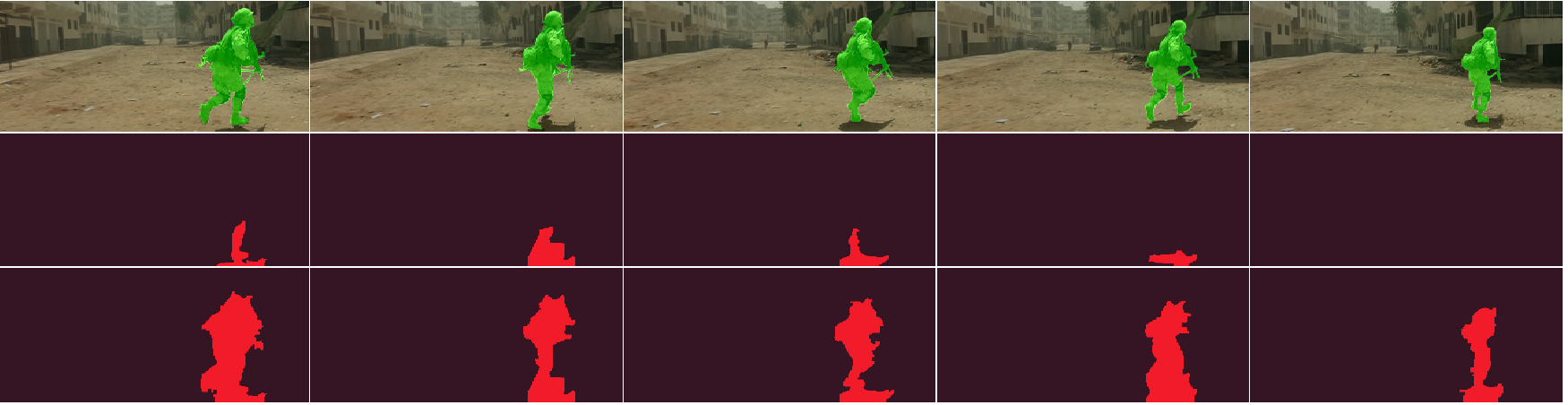}
\end{center}
\caption{The figure shows the segmentation results for the video ``soldier'' from the Segtrack v2 dataset using  spectral clustering \cite{Galasso2013} with and without our feature. We set the number of output clusters at 2 for this example. The top row shows the original frames with the ground truth highlighted in green. The second row shows the results of spectral clustering with 6 features, as originally proposed in \cite{Galasso2013}. The third row shows the results of the algorithm when using the 6 original features plus our feature (2D histogram with spatial information). All other settings were set to be exactly the same.}
\label{fig:soldier}
\end{figure}

An interesting qualitative example is shown in Fig. \ref{fig:soldier}, showing the segmentation results for video ``soldier'' with only two clusters. The second row visualizes the two clusters generated by \cite{Galasso2013} using the 6 predefined features with only local information, only capturing the lower leg of the moving soldier. In contrast, the segmentation results improved with the addition of our geodesic feature. The global information that is encoded by our feature seems to have provided better information to the spectral clustering algorithm to segment the main object out of the background. 
Another qualitative example is shown in the 4th and 5th row of Fig.  \ref{fig:preview}. The segment of the baseline shown in the 4th row shows some under-segmentation over the main moving object. This issue however, is less pronounced with our feature. 

\subsection{Video Segmentation Using Parametric Graph Partitioning.}

\begin{figure}[h]
\begin{center}
\includegraphics[width=\textwidth]{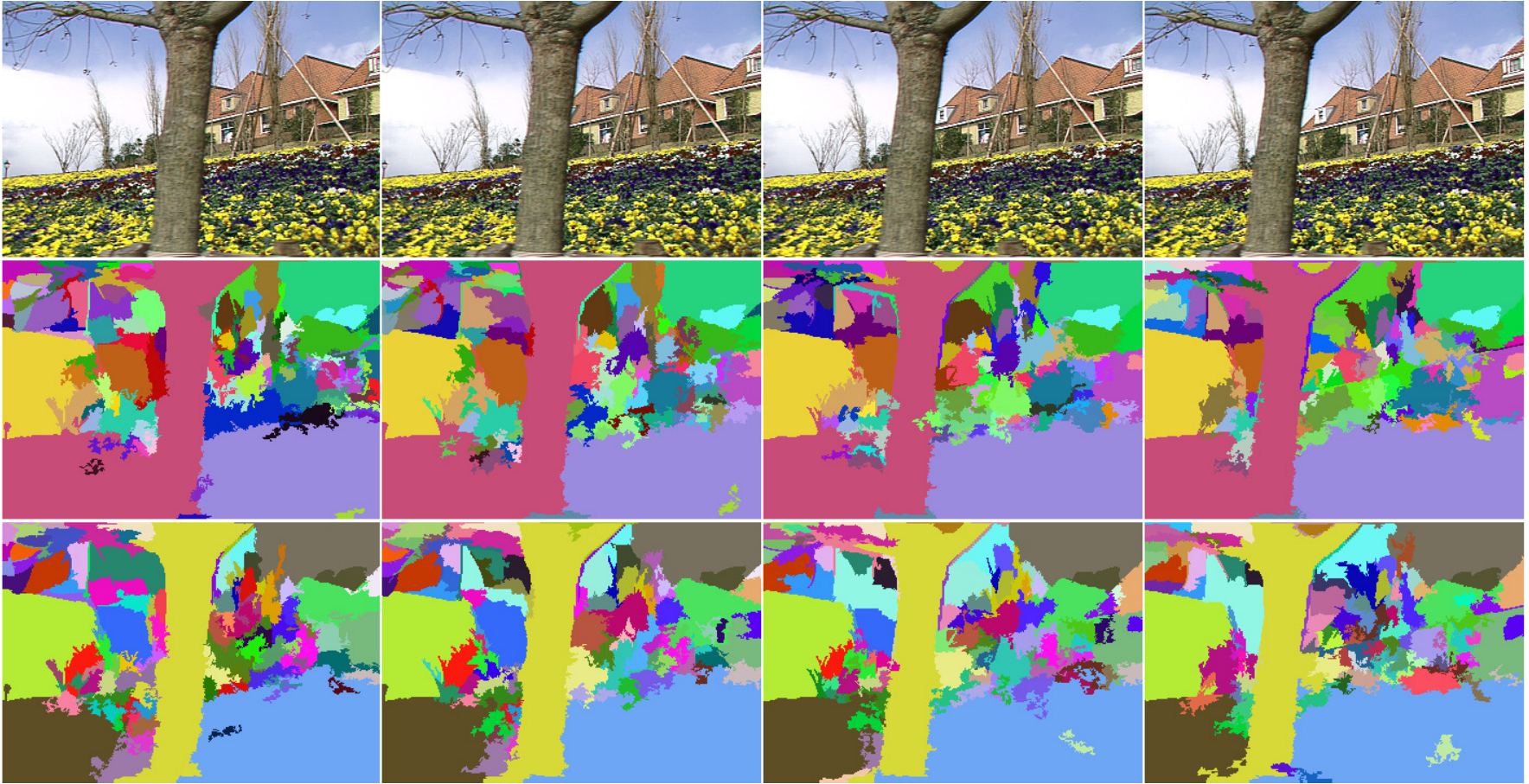}
\end{center}
\caption{The segmentation results of PGP on video ``garden'' from the Chen Xiph.org dataset with and without our feature. The top row shows the original frames. The second row shows the segmentation results of PGP using the 4 features proposed in \cite{Yu_2015_ICCV}. The bottom row is the segmentation results of PGP using the 4 features plus our feature (2D histogram with spatial information).}
\label{fig:cpygd}
\end{figure}

Parametric Graph Partitioning (PGP) \cite{Yu_2015_ICCV} is a recent graph-based unsupervised method that generates a single level of video segmentation. The method models edge weights by a mixture of Weibull distributions, and requires that an $L_p$-norm based similarity distance to be utilized. Therefore, we conduct experiments in this section using Earth Mover's Distance as in \cite{Yu_2015_ICCV}. 
The baseline is the setting originally proposed in \cite{Yu_2015_ICCV} which uses four feature types: intensity, the hue of the HSV color space, the AB component of LAB color space, and gradient orientation. We did not use the motion feature since it did not contribute  significantly toward PGP performance as suggested in the original paper. 

Tables \ref{cptb2} and \ref{cptb1} report the quantitative evaluation of PGP with and without our feature on the two datasets.  
We evaluated the 1D histogram feature on the Chen Xiph.org dataset, shown in Table \ref{cptb1}. While PGP with the 1D feature outperforms the baseline in general, the benchmarks of 3 out of 8 videos decreased. On the other hand, the 2D feature significantly improved the segmentation performance of PGP. For the Segtrack v2 dataset, quantitative results in Table \ref{cptb2} show clear improvements of our feature for PGP, as well as the additional benefits from the spatial pyramid configuration.

Two example cases of PGP are shown in Fig. \ref{fig:cpygd}, and the 2nd and 3rd row of Fig. \ref{fig:preview}. For the over-segmented scenario in Fig.\ref{fig:preview}, the water was unfavorably divided into many spurious segments by the PGP baseline. Adding our feature did not only help merging the background into one segment, but also enhanced temporal consistency and boundary awareness. Given the under-segmented baseline result on the lower part of the tree shown in Fig.\ref{fig:cpygd}, our feature helped to segment the entire tree and also reduced over-segmentation in other parts of the video.

\subsection{Feature Extraction Running Time}

All experiments were conducted on an Intel Core i7 CPU with 3.5 Ghz, and 16 Gb of memory. When adding our feature into the framework of \cite{Galasso2013}, the average additional running time was increased by 67 seconds on a 85-frame video using the default parameter settings, which is a just small fraction of the total running time of several hours. The additional running time increase for the PGP framework was on average 48 seconds, with 300 initial superpixels per frame. These results show that the computational cost of our feature is low, and adds very little overhead to existing frameworks.

\section{Conclusion}
\label{sec:cl}
In this paper, we introduced a novel feature for video segmentation based on geodesic distance histograms. The histogram is computed as a spatially-organized distribution of accumulated boundary costs between superpixels, which is a representation that includes more global information than conventional features. We validated the efficacy of our feature by adding it into two recent frameworks  for video segmentation using spectral clustering and parametric graph partitioning, and showed that the proposed feature improved the performance of both frameworks in 3D video segmentation benchmarks, as well as the temporal consistency of the resulting supervoxels. We believe that the encoded global information can be further applied to other video related tasks such as moving object tracking, object proposals, and foreground background segmentation.

\begin{table}[]
\centering
\caption{Quantitative evaluation on the Chen Xiph.org dataset. Best values are shown in bold. The table shows the evaluation results of the segmentation generated from the method proposed in \cite{Yu_2015_ICCV} with and without our feature in two configurations: 1D geodesic distance histogram and 2D intensity-geodesic distance histogram. All videos are initialized with 300 superpixels.}
\begin{center}
\begin{adjustbox}{max width=\textwidth}
\label{cptb2}
\begin{tabular}{|l|c|c|c|c|c|c|c|c|c|c|c|c|}
\hline
Metrics       & \multicolumn{3}{c|}{3D ACC}                               & \multicolumn{3}{c|}{UE 3D}                                & \multicolumn{3}{c|}{BR 3D}                       & \multicolumn{3}{c|}{BP 3D}                                \\ \hline
Methods       & \cite{Yu_2015_ICCV} & 1D             & 2D             & \cite{Yu_2015_ICCV} & 1D             & 2D             & \cite{Yu_2015_ICCV} & 1D    & 2D             & \cite{Yu_2015_ICCV} & 1D             & 2D             \\ \hline
Bus\_fa       & 70.72                   & 70.58          & \textbf{70.98} & 6.22                    & 10.31          & \textbf{5.75}  & 80.22                   & 81.64 & \textbf{82.46} & 37.64                   & 38.60          & \textbf{38.98} \\ \hline
Container\_fa & 88.68                   & 86.69          & \textbf{89.05} & 3.66                    & 7.54           & \textbf{3.45}  & \textbf{71.24}          & 70.38 & 70.74          & 8.68                    & \textbf{16.28} & 8.55           \\ \hline
Garden\_fa    & 81.69                   & 83.72          & \textbf{85.46} & 1.80                    & 1.68           & \textbf{1.47}  & 72.46                   & 77.48 & \textbf{79.91} & \textbf{12.83}          & 12.73          & 12.41          \\ \hline
Ice\_fa       & 86.71                   & \textbf{87.54} & 77.83          & \textbf{26.70}          & 42.58          & 58.59          & \textbf{83.29}          & 80.82 & 67.47          & 30.99                   & 29.54          & \textbf{44.48} \\ \hline
Paris\_fa     & 40.46                   & 51.37          & \textbf{61.44} & 13.50                   & \textbf{12.99} & 13.15          & 47.17                   & 53.73 & \textbf{56.68} & 4.22                    & 4.70           & \textbf{4.73}  \\ \hline
Soccer\_fa    & 85.79                   & 83.95          & \textbf{87.04} & 4.84                    & 5.46           & \textbf{2.74}  & 31.37                   & 30.47 & \textbf{43.35} & \textbf{5.51}           & 5.20           & 5.49           \\ \hline
Salesman\_fa  & 83.39                   & 72.54          & \textbf{84.69} & 40.48                   & 54.33          & \textbf{12.41} & 73.01                   & 72.76 & \textbf{79.88} & \textbf{22.41}          & 19.93          & 13.47          \\ \hline
Stefan\_fa    & 83.56                   & 81.57          & \textbf{90.14} & 6.76                    & 19.80          & \textbf{4.87}  & 80.66                   & 74.62 & \textbf{83.30} & 10.98                   & \textbf{15.16} & 11.04          \\ \hline
Mean          & 77.62                   & 77.25          & \textbf{80.83} & 12.99                   & 19.34          & \textbf{12.80} & 67.43                   & 67.74 & \textbf{70.47} & 16.66                   & \textbf{17.77} & 17.40          \\ \hline
\end{tabular}
\end{adjustbox}
\end{center}
\end{table}

\begin{table}
\caption{Quantitative evaluation on the SegTrack v2 dataset. Best values are shown in bold. The table shows the evaluation results of the segmentation generated from the algorithm proposed in \cite{Yu_2015_ICCV}, and two of our feature configurations: basic 2D histogram (2D) and 2D histogram with spatial information (2Dsp). The algorithms are all initialized with 300 superpixels per frame.}
\begin{center}

\begin{adjustbox}{max width=\textwidth}
\begin{tabular}{|l|c|c|c||c|c|c||c|c|c||c|c|c|}
\hline
Metrics            & \multicolumn{3}{c|}{3D ACC}                        & \multicolumn{3}{c|}{UE3D}                        & \multicolumn{3}{c|}{BR3D}                        & \multicolumn{3}{c|}{BP3D}                      \\ \hline
Methods            & \cite{Yu_2015_ICCV}              & 2D             & 2Dsp            & \cite{Yu_2015_ICCV}              & 2D             & 2Dsp            & \cite{Yu_2015_ICCV}              & 2D             & 2Dsp            & \cite{Yu_2015_ICCV}              & 2D            & 2Dsp           \\ \hline
B.o.paradise & 96.77          & \textbf{96.81} & 96.79          & \textbf{2.74}  & 3.62           & 3.90           & 93.12          & 94.47          & \textbf{94.80} & 6.83           & \textbf{6.98} & 6.71          \\ \hline
Birdfall           & 58.61          & \textbf{67.54} & 62.22          & 24.42          & 11.15          & \textbf{10.52} & 77.99          & 90.92          & \textbf{92.36} & \textbf{0.61}  & 0.45          & 0.47          \\ \hline
Bmx-1              & \textbf{94.60} & 94.50          & 94.56          & 5.49           & 6.44           & \textbf{5.40}  & 98.31          & 98.32          & \textbf{98.58} & 4.05           & \textbf{4.66} & 4.23          \\ \hline
Bmx-2              & 78.00          & 78.39          & \textbf{81.40} & \textbf{11.43} & 13.33          & 12.48          & 94.00          & 91.49          & \textbf{95.01} & 3.72           & \textbf{4.17} & 3.92          \\ \hline
Cheetah-1          & 73.26          & 75.76          & \textbf{76.35} & 30.62          & 6.59           & \textbf{5.46}  & 92.19          & 97.54          & \textbf{98.62} & \textbf{1.65}  & 1.09          & 1.10          \\ \hline
cheetah-2          & 63.84          & \textbf{73.68} & 69.38          & 34.64          & \textbf{6.95}  & 8.73           & 97.85          & 98.54          & \textbf{98.66} & \textbf{2.19}  & 1.38          & 1.38          \\ \hline
Drift-1            & \textbf{93.85} & 93.20          & 93.34          & 3.77           & \textbf{3.29}  & 3.42           & 92.70          & \textbf{94.54} & 94.53          & \textbf{1.22}  & 1.20          & \textbf{1.22} \\ \hline
Drift-2             & \textbf{92.43} & 92.41          & 92.06          & 3.31           & 2.98           & \textbf{2.96}  & 90.52          & \textbf{92.53} & 92.13          & \textbf{0.94}  & 0.93          & \textbf{0.94} \\ \hline
Frog               & 56.92          & 64.72          & \textbf{86.67} & 16.32          & 14.01          & \textbf{11.60} & 59.28          & 76.14          & \textbf{83.26} & \textbf{10.42} & 3.84          & 2.25          \\ \hline
Girl               & 87.71          & \textbf{89.18} & \textbf{89.18} & 10.76          & \textbf{10.18} & 10.27          & 90.18          & 94.59          & \textbf{94.68} & \textbf{5.46}  & 5.39          & 5.32          \\ \hline
Hum.bird-1      & 65.07          & \textbf{73.32} & 73.27          & 9.41           & \textbf{9.16}  & 9.20           & \textbf{88.50} & 88.48          &87.10 & 3.14           & 3.26          & \textbf{3.76} \\ \hline
Hum.bird-2      & 77.71          & 84.95          & \textbf{85.52} & \textbf{6.35}  & 7.04           & 9.06           & \textbf{94.64} & 94.26          & 94.58          & 5.00           & 5.18          & \textbf{6.09} \\ \hline
Monkey             & 86.86          & 89.06          & \textbf{89.62} & 13.66          & 3.84           & \textbf{3.73}  & 93.07          & 98.32          & \textbf{98.37} & \textbf{2.79}  & 1.59          & 1.62          \\ \hline
M.dog-1        & 88.09          & 88.70          & \textbf{88.97} & 9.50           & \textbf{9.30}  & 9.38           & 95.74          & 97.44          & \textbf{98.50} & 1.40           & \textbf{1.42} & \textbf{1.42} \\ \hline
M.dog-2        & 62.57          & \textbf{65.60} & 64.79          & 5.82           & 5.36           & \textbf{5.15}  & 86.80          & \textbf{91.13} & 90.56          & 0.91           & \textbf{0.95} & 0.94          \\ \hline
Parachute          & 92.54          & 92.31          & \textbf{92.31} & 19.54          & 18.29          & \textbf{5.65}  & 95.24          & 95.71          & \textbf{97.34} & \textbf{1.27}  & 1.13          & 0.76          \\ \hline
Penguin-1          & \textbf{95.72} & 23.36          & 93.45          & 3.38           & \textbf{3.56}  & \textbf{3.56}  & \textbf{49.25} & 44.57          & 44.53          & \textbf{0.89}  & 0.83          & 0.66          \\ \hline
Penguin-2          & 95.51          & 95.77          & \textbf{95.79} & 3.39           & \textbf{3.28}  & \textbf{3.28}  & 73.19          & 71.41          & \textbf{74.78} & 1.38           & \textbf{1.39} & 1.17          \\ \hline
Penguin-3          & 96.49          & \textbf{96.79} & 96.48          & 3.87           & 3.87           & \textbf{3.83}  & 67.89          & 68.13          & 74.44          & 1.28           & \textbf{1.32} & 1.16          \\ \hline
Penguin-4          & \textbf{95.72} & 94.50          & 94.74          & \textbf{3.87}  & 3.95           & 3.92           & 73.54          & 73.82          & \textbf{73.44} & 1.16           & \textbf{1.21} & 0.96          \\ \hline
Penguin-5          & \textbf{93.27} & 92.25          & 91.63          & 8.38           & \textbf{8.21}  & 8.22           & \textbf{74.01} & 72.87          & 71.14          & 1.03           & \textbf{1.05} & 0.82          \\ \hline
Penguin-6          & 92.37          & 92.64          & \textbf{93.09} & \textbf{3.73}  & 4.02           & 4.03           & 62.33          & \textbf{63.52} & 59.50          & 1.02           & \textbf{1.08} & 0.81          \\ \hline
Soldier            & 89.81          & \textbf{90.19} & \textbf{90.19} & 4.71           & \textbf{4.11}  & 4.40           & 92.38          & 93.29          & \textbf{93.48} & 1.87           & 1.86          & \textbf{1.89} \\ \hline
Worm               & 92.21          & 92.71          & \textbf{92.75} & \textbf{10.31} & 15.18          & 14.95          & 89.28          & 92.72          & \textbf{93.48} & 1.01           & \textbf{1.19} & 1.17          \\ \hline
\textbf{Average}   & 84.16          & 83.26          & \textbf{86.86} & 10.39          & 7.40           & \textbf{6.80}  & 84.25          & 86.45          & \textbf{87.24} & \textbf{2.55}  & 2.23          & 2.12          \\ \hline
\end{tabular}
\end{adjustbox}
\end{center}
\label{cptb1}
\end{table}

\noindent {\bf Acknowledgement}. Partially supported by the Vietnam Education Foundation, NSF IIS-1161876, FRA DTFR5315C00011,  the Stony Brook SensonCAT, the SubSample project from the DIGITEO Institute, France, and a gift from Adobe Corporation
\FloatBarrier

\vspace{3mm}

\bibliographystyle{splncs}
\bibliography{0145}


\end{document}